\documentclass[10pt,twocolumn,twoside]{article}

\usepackage[utf8]{inputenc}

\usepackage{amssymb,amsmath,amsthm,mathtools,graphicx} 
\usepackage{bm} 
\usepackage{cases} 
\usepackage{diffcoeff} 
\usepackage{booktabs} 
\usepackage{array} 
\usepackage{tabularx} 
\usepackage{makecell} 
\usepackage{collcell} 
\usepackage{multirow} 
\usepackage[table]{xcolor} 
\usepackage{etoolbox,siunitx} 
\robustify\bfseries
\usepackage{float} 
\usepackage{color} 
\usepackage{textcomp} 
\usepackage{marvosym, utfsym, gensymb} 
\robustify\usym
\usepackage{adjustbox} 
\usepackage{comment} 
\usepackage{censor} 
\usepackage{import} 
\usepackage[inline]{enumitem} 
\usepackage{refcount} 
\usepackage{algorithm}
\usepackage{algpseudocode}
\usepackage{xifthen} 
\usepackage{xparse} 
\usepackage{xstring} 
\usepackage[numbers,sort]{natbib} 
\usepackage[pdfusetitle,hidelinks,plainpages=false]{hyperref}
\usepackage{doi} 
\NewDocumentCommand\ProcStar{m}{
    \IfBooleanTF{#1}{%
        \def\ProcessedArgument{*}%
    }{%
        \def\ProcessedArgument{}%
    }%
}
\makeatletter
\NewDocumentCommand\subtoarg{m m}{
    \expandafter\def\csname #1@sub\endcsname_##1{\csname #2\endcsname[##1]}
    \expandafter\def\csname #1\endcsname{
        \@ifnextchar_{\csname #1@sub\endcsname}{\csname #2\endcsname[]}
    }%
}
\newlength{\boxwidth}
\newlength{\boxheight}
\newsavebox{\boxdim@box}
\NewDocumentCommand\boxdim{m}{
    \sbox{\boxdim@box}{#1}%
    \setlength{\boxwidth}{\wd\boxdim@box}%
    \setlength{\boxheight}{\dimexpr\ht\boxdim@box + \dp\boxdim@box\relax}%
}
\newsavebox{\autoparbox@box}
\NewDocumentCommand\autoparbox{O{} O{} O{} m}{%
    \sbox{\autoparbox@box}{#4}%
    \parbox[#1][#2][#3]{\wd\autoparbox@box}{\usebox{\autoparbox@box}}%
}
\makeatother
\providecommand\given{} 
\newcommand\GivenResize[1][]{
    \RenewDocumentCommand\given{t~}{\nonscript\:#1\vert\IfBooleanF{##1}{\allowbreak}\nonscript\:\mathopen{}}}
\GivenResize[]
\DeclarePairedDelimiter\abs{\lvert}{\rvert} 
\DeclarePairedDelimiter\ceil{\lceil}{\rceil} 
\DeclarePairedDelimiter\floor{\lfloor}{\rfloor} 
\DeclarePairedDelimiter\parens{(}{)} 
\DeclarePairedDelimiter\bracks{[}{]} 
\DeclarePairedDelimiter\tuple{(}{)} 
\DeclarePairedDelimiterX\set[1]{\lbrace}{\rbrace}{\GivenResize[\delimsize]#1} 
\DeclarePairedDelimiterX\group[2]{\langle}{\rangle}{\GivenResize[\delimsize]#1} 
\DeclarePairedDelimiterX\civ[2]{[}{]}{#1\,,\,\allowbreak#2} 
\DeclarePairedDelimiterX\oiv[2]{(}{)}{#1\,,\allowbreak#2} 
\DeclarePairedDelimiterX\coiv[2]{[}{)}{#1\,,\allowbreak#2} 
\DeclarePairedDelimiterX\ociv[2]{(}{]}{#1\,,\allowbreak#2} 
\DeclarePairedDelimiterXPP\prob[1]{P}{(}{)}{}{\GivenResize[\delimsize]#1} 
\makeatletter
\NewDocumentCommand\smashop{O{btlr} m}{
    \def\smashop@vargs{}
    \def\smashop@hargs{}
    \IfSubStr{#1}{t}{\g@addto@macro\smashop@vargs{t}}{}
    \IfSubStr{#1}{b}{\g@addto@macro\smashop@vargs{b}}{}%
    \IfSubStr{#1}{l}{\g@addto@macro\smashop@hargs{l}}{}%
    \IfSubStr{#1}{r}{\g@addto@macro\smashop@hargs{r}}{}%
    \smash[\smashop@vargs]{%
        \ifthenelse{\isempty{\smashop@hargs}}{%
            #2%
        }{%
            \smashoperator[\smashop@hargs]{#2}%
        }%
    }%
}
\newcommand\ExpectSymbol{\mathbb{E}}
\newcommand\ExpectMargin{1pt 0}
\DeclareMathOperator*{\ExpectOp}{\mathchoice
    {\vcenter{\hbox{\marginbox{\ExpectMargin}{\resizebox{10pt}{!}{$\ExpectSymbol$}}}}}
    {\ExpectSymbol}
    {\ExpectSymbol}
    {\ExpectSymbol}}
\DeclarePairedDelimiterX\ExpectDelim[1]{[}{]}{\GivenResize[\delimsize]#1}
\NewDocumentCommand\expect{o e{_} s o g}{
    \IfValueTF{#2}{
        \def\expect@size{#4}%
    }{
        \def\expect@size{#1}%
    }%
    \ExpandArgs{o}\IfValueTF{\expect@size}{
        \edef\expect@size{[\expandafter\noexpand\expect@size]}
    }{
        \def\expect@size{}%
    }%
    \IfBooleanT{#3}{
        \def\expect@size{*}%
    }%
    \IfValueTF{#2}{
        \IfValueTF{#1}{
            \smashop[#1]{\ExpectOp_{#2}}%
        }{
            \ExpectOp_{#2}%
        }%
    }{
        \ExpectOp%
    }%
    \IfValueT{#5}{
        \expandafter\ExpectDelim\expect@size{#5}
    }%
}
\DeclareMathOperator*{\entropyOp}{\mathrm{H}}
\DeclarePairedDelimiterXPP\entropyDelim[1]{\entropyOp}{(}{)}{}{#1}
\NewDocumentCommand\entropy{>{\ProcStar}s O{} m g}{\entropyDelim#1[#2]{#3\IfValueT{#4}{,#4}}}
\DeclareMathOperator*{\KLOp}{\mathrm{KL}}
\DeclarePairedDelimiterXPP\KL[2]{\KLOp}{[}{]}{}{\GivenResize[\delimsize\vert\delimsize]#1 \given #2}
\newcommand\union{
    \@ifnextchar_{%
        \bigcup%
    }{%
        \@ifnextchar^{%
            \bigcup%
        }{%
            \cup%
        }%
    }%
}
\robustify\union
\newcommand\intersec{
    \@ifnextchar_{%
        \bigcap%
    }{%
        \@ifnextchar^{%
            \bigcap%
        }{%
            \cap%
        }%
    }%
}
\robustify\intersec
\makeatother
\newcommand\mat[1]{\begin{bmatrix}#1\end{bmatrix}} 
\DeclareMathOperator*{\argmax}{arg\,max}

\def\mathscale{1}
\NewDocumentCommand\scalemath{O{\mathscale} m}{\scalebox{#1}{\mbox{\ensuremath{\displaystyle #2}}}}

\makeatletter
\let\mss@ifnextchar\@ifnextchar
\NewDocumentCommand\mss{e{_} E{'}{{}} e{^}}{
    \def\mss@sub{#1}
    \def\mss@prime{#2}
    \def\mss@sup{#3}
    \mss@ifnextchar_{
        \mss@submerge
    }{%
        \mss@ifnextchar^{
            \mss@supmerge
        }{%
            \mss@ifnextchar'{
                \mss@primemerge
            }{
                \IfValueT{#1}{_{#1}}%
                \IfValueTF{#3}{#2^{#3}}{#2}
            }
        }%
    }%
}%
\NewDocumentCommand\mss@merge{m m}{
    \IfNoValueTF{#1}{
        #2%
    }{%
        \IfNoValueTF{#2}{
            #1%
        }{
            \ifthenelse{\isempty{#1}}{
                #2%
            }{%
                \ifthenelse{\isempty{#2}}{
                    #1%
                }{%
                    #1,#2%
                }%
            }%
        }%
    }%
}
\def\mss@submerge_#1{
    \edef\mss@sub{\mss@merge{\expandonce{\mss@sub}}{\unexpanded{#1}}}%
    \edef\mss@args{_{\expandonce{\mss@sub}}'{\expandonce{\mss@prime}}^{\expandonce{\mss@sup}}}%
    \expandafter\mss\mss@args
}
\def\mss@supmerge^#1{
    \edef\mss@sup{\mss@merge{\expandonce{\mss@sup}}{\unexpanded{#1}}}%
    \edef\mss@args{_{\expandonce{\mss@sub}}'{\expandonce{\mss@prime}}^{\expandonce{\mss@sup}}}%
    \expandafter\mss\mss@args
}
\def\mss@primemerge'{
    \edef\mss@prime{\expandonce{\mss@prime}'}%
    \edef\mss@args{_{\expandonce{\mss@sub}}'{\expandonce{\mss@prime}}^{\expandonce{\mss@sup}}}%
    \expandafter\mss\mss@args
}
\robustify\mss 
\makeatother

\newcommand\lb[1]{{#1}\mss_{\textsc{l}}^{}} 
\newcommand\ub[1]{{#1}\mss_{\textsc{u}}^{}} 
\newcommand\ind{I} 

\makeatletter
\newcommand{\newreptheorem}[2]{%
    \newtheorem{rep@#1}{#2}%
    \NewDocumentEnvironment{rep#1}{m o}{%
        \lsuffix@set{##1}[##2]%
        \setcounterref{rep@#1}{\lsuffix@label}%
        \addtocounter{rep@#1}{-1}%
        \begin{rep@#1}%
    }{%
        \end{rep@#1}%
    }%
}
\makeatother

\newreptheorem{theorem}{Theorem}

\newreptheorem{lemma}{Lemma}

\newreptheorem{proposition}{Proposition}

\newtheorem*{remark}{Remark} 

\newcommand\mail[1]{\Letter\,\href{mailto:#1}{#1}}
\NewDocumentCommand\AuthorCite{o m}{%
    \IfNoValueTF{#1}{
        \begin{NoHyper}\citeauthor{#2}\end{NoHyper}~\cite{#2}%
    }{
        \begin{NoHyper}\citeauthor{#1}\end{NoHyper}~\cite{#1,#2}%
    }%
}

\newcommand\sq[1]{`#1'} 

\AtBeginDocument{
  \let\oldlabel\label
  \let\oldref\ref
  \let\label\suffixedlabel
  \let\ref\suffixedref
}

\makeatletter
\def\lsuffix@sep{-}
\def\lsuffix@root{root}
\def\lsuffix@old{!}
\def\lsuffix{\lsuffix@root}
\NewDocumentCommand\lsuffix@set{m o}{%
  \IfNoValueTF{#2}{
    \def\lsuffix@suf{\lsuffix}%
  }{\ifthenelse{\isempty{#2}}{
    \def\lsuffix@suf{\lsuffix@root}%
  }{
    \def\lsuffix@suf{#2}%
  }}%
  \IfStrEq{\lsuffix@old}{\lsuffix@suf}{%
    \def\lsuffix@label{#1}
  }{%
    \def\lsuffix@label{#1\lsuffix@sep\lsuffix@suf}
  }%
}
\NewDocumentCommand\suffixedlabel{m o}{%
  \lsuffix@set{#1}[#2]%
  \expandafter\oldlabel\expandafter{\lsuffix@label}%
}
\NewDocumentCommand\suffixedref{m o}{%
  \lsuffix@set{#1}[#2]%
  \expandafter\oldref\expandafter{\lsuffix@label}%
}
\newcommand{\setlabelsuffix}[1]{\def\lsuffix{#1}}
\newcommand{\unsetlabelsuffix}{\def\lsuffix{\lsuffix@root}}
\makeatother

\NewDocumentCommand\chapref{m o}{Chapter~\ref{#1}[#2]}
\NewDocumentCommand\appref{m o}{Appendix~\ref{#1}[#2]}
\NewDocumentCommand\secref{m o}{Section~\ref{#1}[#2]}
\NewDocumentCommand\ssecref{m o}{Subsection~\ref{#1}[#2]}
\NewDocumentCommand\figref{m o}{Figure~\ref{#1}[#2]}
\NewDocumentCommand\tabref{m o}{Table~\ref{#1}[#2]}
\NewDocumentCommand\thmref{m o}{Theorem~\ref{#1}[#2]}
\NewDocumentCommand\lemref{m o}{Lemma~\ref{#1}[#2]}
\NewDocumentCommand\corref{m o}{Corollary~\ref{#1}[#2]}
\NewDocumentCommand\propref{m o}{Proposition~\ref{#1}[#2]}
\let\oldalgref\algref
\RenewDocumentCommand\algref{m g o}{Algorithm~\IfValueTF{#2}{\oldalgref{#1}{#2}}{\ref{#1}[#3]}}
\makeatletter
\RenewDocumentCommand\eqref{m}{\textup{\tagform@{\oldref{#1}}}}
\makeatother

\newboolean{blinded}
\setboolean{blinded}{false}
\NewDocumentCommand\blind{m o}{%
    \ifthenelse{\boolean{blinded}}{
        \IfNoValueTF{#2}{
            \xblackout{#1}%
        }{
            \xblackout{#2}%
        }%
    }{
        #1%
    }%
}
\NewDocumentCommand\blindfootnotehref{m m}{%
    \ifthenelse{\boolean{blinded}}{
        #2\footnote{\blind{#1}}%
    }{
        \href{#1}{#2}\footnote{\url{#1}}%
    }%
}

\NewDocumentCommand\timesub{m}{%
    t%
    \IfInteger{#1}{%
        \ifnum0<0#1
            +#1%
        \else\ifnum0=#1
        \else
            #1%
        \fi\fi%
    }{
        +#1%
    }%
}
\NewDocumentCommand\namesub{m}{%
    \textsc{\lowercase{#1}}%
}
\NewDocumentCommand\mntsub{g o}{%
    \IfValueTF{#1}{%
        \IfValueTF{#2}{%
            \mss_{\namesub{#1}}_{\timesub{#2}}%
        }{%
            \mss_{\namesub{#1}}%
        }%
    }{%
        \IfValueTF{#2}{%
            \mss_{\timesub{#2}}%
        }{%
            \mss_{}%
        }%
    }%
}
\NewDocumentCommand\s{o}{
    \bm{s}\mntsub[#1]^{}
}
\RenewDocumentCommand\a{g o}{
    \bm{a}\mntsub{#1}[#2]^{}%
}
\NewDocumentCommand\an{o}{
    \tilde{\bm{a}}\mntsub[#1]^{}%
}
\RenewDocumentCommand\o{g o}{
    o\mntsub{#1}[#2]^{}%
}
\renewcommand\S{\mathcal{S}}
\NewDocumentCommand\A{g}{
    \mathcal{A}\mntsub{#1}^{}%
}
\newcommand\p{\bm{\theta}}
\newcommand\exptup{\tuple{\s[0],\a[0],r_t,\s[1]}}
\newcommand\buf{\mathcal{B}}
\NewDocumentCommand\pol{g}{\pi\mntsub{#1}^{}}
\NewDocumentCommand\bpol{g}{\beta\mntsub{#1}^{}}
\RenewDocumentCommand\O{g}{\mathcal{O}\mntsub{#1}^{}}
\NewDocumentCommand\Oav{g}{\O{#1}^{\textsc{av}}}
\newcommand\I{\mathcal{I}}
\newcommand\T{\mathcal{T}}
\NewDocumentCommand\gauss{}{\mathcal{N}}
\NewDocumentCommand\uniform{m}{\mathrm{U}\bracks{#1}}
\newcommand\rw[1]{w\mntsub{#1}^{}}
\newcommand\rc[1]{r\mntsub{#1}^{}}
\newcommand\vt{v_\mathrm{T}^{}} 
\newcommand\dt{d_\mathrm{T}^{}} 

\usepackage[margin=1.6cm]{geometry}
\date{}
\let\oldmaketitle\maketitle
\renewcommand\maketitle{%
    \begingroup%
    \def\thefootnote{}%
    \def\footnotemark{}%
    \oldmaketitle%
    \endgroup%
}
\let\oldtabular\tabular%
\renewcommand\tabular{\footnotesize\oldtabular}%
\providecommand\initenvtitle{}
\newtheorem*{initenv}{\small\initenvtitle}
\RenewDocumentEnvironment{abstract}{}{\renewcommand\initenvtitle{Abstract}\initenv\small}{\endinitenv}
\NewDocumentEnvironment{keywords}{}{\renewcommand\initenvtitle{Keywords}\initenv\small}{\endinitenv}
\usepackage{caption}
\providecommand\appendices{%
    \gdef\thesection{\Alph{section}}%
    \setcounter{section}{0}%
    \setcounter{subsection}{0}%
    \setcounter{subsubsection}{0}%
    \setcounter{paragraph}{0}%
    \section*{Appendices}%
}

\newcommand\orcidID[1]{%
}
\newcommand\corAuth{\textsuperscript{(\Letter)}}

\title{Learning to Drive Safely with Hybrid Options}
\author{Bram De Cooman\orcidID{0000-0003-4843-3342}\corAuth,
Johan Suykens\orcidID{0000-0002-8846-6352}
\thanks{B. {De Cooman} (\mail{bram.decooman@esat.kuleuven.be}) and J. Suykens (\mail{johan.suykens@esat.kuleuven.be}) are with STADIUS Center for Dynamical Systems, Signal Processing and Data Analytics, Department of Electrical Engineering (ESAT), KU Leuven, 3001 Leuven, Belgium.}%
}
\begin{document}

\maketitle

\begin{abstract}
    Out of the many deep reinforcement learning approaches for autonomous driving, only few make use of the options (or skills) framework. That is surprising, as this framework is naturally suited for hierarchical control applications in general, and autonomous driving tasks in specific.
    Therefore, in this work the options framework is applied and tailored to autonomous driving tasks on highways. More specifically, we define dedicated options for longitudinal and lateral manoeuvres with embedded safety and comfort constraints. This way, prior domain knowledge can be incorporated into the learning process and the learned driving behaviour can be constrained more easily.
    We propose several setups for hierarchical control with options and derive practical algorithms following state-of-the-art reinforcement learning techniques. By separately selecting actions for longitudinal and lateral control, the introduced policies over combined and hybrid options obtain the same expressiveness and flexibility that human drivers have, while being easier to interpret than classical policies over continuous actions.
    Of all the investigated approaches, these flexible policies over hybrid options perform the best under varying traffic conditions, outperforming the baseline policies over actions.
\end{abstract}

\section{Introduction}
Due to its great potential to solve complex decision-making problems, Reinforcement Learning (RL) is a promising technique to train virtual drivers for autonomous driving applications~\cite{Kiran2021,Naveed2021,Aradi2022,Wang2023,Zhou2023,Gurses2024,Li2024}. An important design decision is the definition of the action space, defining how the virtual driver (agent) interacts with the environment. Most attempts so far have restricted their action spaces to be either fully discrete (with actions such as \sq{lane change left} and \sq{brake})~\cite{Nageshrao2019,Alizadeh2019} or fully continuous (with actions such as \sq{turn the steering wheel by $x\degree$} or \sq{accelerate by $y\,\unit{m/s^2}$})~\cite{Saxena2020,Hou2025}. In this paper, we investigate some alternative approaches with options. The resulting hierarchical control architectures using combined or hybrid options have several benefits over the fully discrete or continuous approaches.

A classical RL setup with a discrete action space requires each action to have the same duration, corresponding to exactly one timestep in the underlying Markov Decision Process (MDP). An MDP with options, on the other hand, inherently supports different durations for each option. This makes the options approach especially useful for autonomous driving applications, in which a virtual driver needs access to both fast acceleration manoeuvres and slower lane change manoeuvres.

A drawback of purely discrete action spaces is that they lack flexibility and expressiveness, as, for example, the velocity of the ego vehicle can only be changed in fixed increments. This can be resolved using continuous action spaces. However, the extra flexibility of continuous actions also comes at a cost. Typically, the training time significantly increases and it becomes much harder to constrain the learned policies. Extra penalties could be introduced in the reward signal to try and constrain the vehicle's driving behaviour. However, this approach becomes increasingly more complex as the number of constraints grows. Moreover, determining good penalties for manoeuvres that take multiple timesteps to complete, is not a straightforward task either. In contrast, under the introduced options framework, we have the ability to constrain vehicle movements more reliably using dedicated options with embedded safety and comfort constraints for certain manoeuvres.

The proposed framework for hierarchical control with options has several other benefits. It has the ability to flexibly combine dedicated options for longitudinal and lateral control, for which already existing driving assistance systems (such as cruise control or lane assist) can be reused. This can accelerate the learning process, as individual driving manoeuvres no longer have to be learned from scratch. Additionally, it can improve the trust of the general public in the learned driving policies. In particular, the vehicle will never perform any other (weird) manoeuvres than the predefined ones, which can be based on automated features that are already widely used and trusted by human drivers. Finally, the obtained models are also more interpretable than their continuous counterparts, as it is immediately clear what kind of manoeuvre the ego vehicle is performing by looking at which option is active.
Ultimately, the proposed hybrid policies with options share the same expressiveness and flexibility as human drivers, effectively combining the best of discrete and continuous approaches.

\subsection*{Contributions}
There are two key contributions in this work. First, we propose a set of options and associated manoeuvres tailored to autonomous driving tasks. We show how safety measures applied to policies over continuous actions can be reapplied to the hierarchical control setup with options, resulting in safe and comfortable driving behaviour by design. Second, we introduce two novel hierarchical control architectures with combined options and hybrid options. Both support the separate selection of actions for longitudinal and lateral control, leading to more flexible driving policies.

\subsection*{Organization}
This paper is further organized as follows. \secref{sec:prelim} gives an introduction to reinforcement learning, the options framework, and the methods used throughout this work. The autonomous driving setup and simulation environment are briefly reviewed in \secref{sec:ads}. Afterwards, in \secref{sec:methods} the practical algorithms for hierarchical control with options tailored to the autonomous driving task are introduced. The proposed methodology is evaluated and compared with other approaches in \secref{sec:experiments}. Finally, the related work is discussed in \secref{sec:related} before the conclusion.

\section{Preliminaries}\label{sec:prelim}
The basic concepts of reinforcement learning and the options framework are briefly introduced here. For a more detailed overview, the reader is referred to Sutton et al. \cite{Sutton1999,Sutton2018}.

\subsection{Reinforcement Learning}
Reinforcement Learning (RL) is a machine learning technique that can be applied to solve sequential decision-making problems, in which an agent interacts with an environment.

\subsubsection*{Markov Decision Process}
Formally, the environment can be described as a Markov Decision Process (MDP) \cite{Puterman1994} $\mathcal{M} = \tuple{\S, \A, \iota, \tau, r, \gamma}$ consisting of state space $\S$, action space $\A$, initial state distribution $\iota$, state transition distribution $\tau$, reward function $r$ and discount factor $\gamma$.
At every timestep $t$, the agent is able to observe the environment's current state $\s[0] \in \S$ and perform a certain action $\a[0] \in \A$.
This brings the environment to a new state $\s[1]$ in the following timestep, according to the stochastic environment dynamics $\tau(\s[1] \given \s[0], \a[0])$.
Simultaneously, the agent receives a scalar reward signal $r_t = r(\s[0], \a[0], \s[1])$, indicating how favorable the taken action and resulting state transition are.
This process of observing a state (and reward) and executing an action is continuously repeated throughout an episode, and is started by sampling an initial state from $\iota(\s_0)$.

\subsubsection*{Policy and Objective}
The agent's policy $\pi(\a[0] \given \s[0])$ describes the probability (density) of taking action $\a[0]$ after perceiving $\s[0]$.
This policy can be improved by taking feedback from the reward signal into account.
More specifically, the goal in reinforcement learning is to find the optimal policy $\pi^*$, maximizing the expected discounted return
\begin{equation}
    \pi^* = \argmax_\pi \expect_{\iota, \pi, \tau}*{\sum_{t=0}^{\infty} \gamma^t r(\s[0], \a[0], \s[1])}.
    \label{eq:rl_objective}
\end{equation}
The expectation is taken over $\s_0 \sim \iota(\cdot)$, $\a[0] \sim \pi(\cdot \given \s[0])$ and $\s[1] \sim \tau(\cdot \given \s[0], \a[0])$ for all timesteps.
The discount factor $\gamma \in \coiv{0}{1}$ ensures the discounted return objective remains finite for bounded reward functions and trades off the importance of future versus immediate rewards.
For deterministic policies, which we primarily work with in this paper, the notation $\a = \pi(\s)$ is used.

\subsubsection*{Value Function}
It is often helpful to consider the value functions of a certain policy $\pi$, defined as $v^\pi(\s) = \expect_{\pi, \tau}{\sum_{k=0}^{\infty} \gamma^k r(\s[k], \a[k], \s[k+1]) \given {\s[0]=\s}}$ and $q^\pi(\s, \a) = \expect_{\pi, \tau}{\sum_{k=0}^{\infty} \gamma^k r(\s[k], \a[k], \s[k+1]) \given {\s[0]=\s},\, {\a[0]=\a}}$.
These value functions aid in evaluating the respective policy, as they both provide the expected discounted return when following policy $\pi$ starting from an arbitrary state $\s$ (and action $\a$ for $q^\pi$).
Furthermore, the value functions for the optimal policy $\pi^*$ satisfy the Bellman optimality equations $v^*(\s) = \max_{\a} \expect_{\tau}{r(\s[0], \a[0], \s[1]) + \gamma v^*(\s[1]) \given {\s[0]=\s},\, {\a[0]=\a}}$ and $q^*(\s, \a) = \expect_{\tau}{r(\s[0], \a[0], \s[1]) + \gamma \max_{\a'} q^*(\s[1], \a') \given {\s[0]=\s},\,\allowbreak {\a[0]=\a}}$.

\subsubsection*{Parameterized Actions}
Although the action space of an MDP is usually either continuous or discrete, more complex environments with hybrid action spaces can also be considered. In a \emph{Parameterized Action MDP} (PAMDP)~\cite{Masson2016}, the hybrid action space has a particular structure, consisting of a discrete set of actions $\A{d} = \bracks{D} = \set{1, \dots, D}$, each parameterized by a continuous set of parameters $\A{c}_{\a{d}} \subseteq \mathbb{R}^m$. More precisely, the parameterized action space can be defined as $\A = \set{(\a{d}, \a{c}) \given \a{d} \in \A{d}, \a{c} \in \A{c}_{\a{d}}}$. For such PAMDPs, the policy can be decomposed into a discrete and continuous part as $\pi(\a \given \s) = \pol{d}(\a{d} \given \s) \,\pol{c}_{\a{d}}(\a{c} \given \s)$. Sampling an action $\a = \tuple{\a{d}, \a{c}}$ can then be done in two steps: first the discrete action $\a{d}$ is sampled from $\pol{d}$, and afterwards the continuous parameters $\a{c}$ are sampled from the corresponding $\pol{c}_{\a{d}}$.

\subsubsection*{Model-Free RL and Exploration}
In the model-free RL setting used throughout this work, the environment dynamics and reward function ($\iota$, $\tau$, $r$) remain unknown to the agent.
To find an optimal policy, the agent is thus required to explore the state-action space during training.
For this purpose, a behavioural policy $\bpol(\a \given \s)$ is deployed, for which the encountered experience tuples $\exptup$ are collected in a replay buffer $\buf$.
Batches of experience tuples sampled from this replay buffer can then be used for optimizing the policy and other models.
To sufficiently explore the state-action space when working with deterministic policies, $\epsilon$-greedy behavioural policies are typically used for discrete action spaces, and Gaussian behavioural policies for continuous action spaces.
Such an $\epsilon$-greedy policy takes a random action with probability $\epsilon$ and the greedy action following the policy $\pi(\s)$ otherwise, i.e. $\bpol(\a \given \s) = \epsilon \abs{\A}^{-1} + (1 - \epsilon) \ind_{\set{\pi(\s)}}(\a)$ where $\ind_{\mathcal{X}}(x)$ is the indicator function --- equal to $1$ when $x \in \mathcal{X}$ and $0$ otherwise.
The Gaussian policy, on the other hand, has the action taken by $\pi$ as mean and $\sigma^2$ as variance, i.e. $\a \sim \gauss(\pi(\s), \sigma^2 I)$.
The $\epsilon$ and $\sigma$ parameters trade off the amount of exploration versus exploitation of the learned policy, and are typically attenuated throughout the training process.

\subsection{Options}
\begin{figure}
    \centering
    \includegraphics[width=\columnwidth]{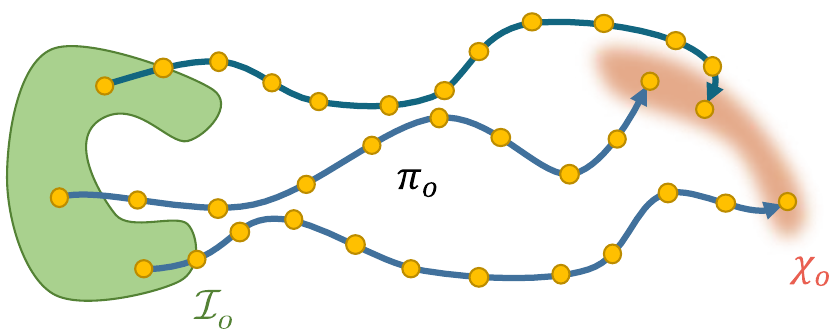}
    \caption{Schematic overview of an option acting in state space. Once in control, the option policy $\pi_{\o}$ can bring the agent from states in the initiation set $\I_{\o}$ all the way to states with nonzero termination probabilities $\chi_{\o}$. For each intermediate state (yellow dots), the option policy provides the environment with suitable actions at every timestep to realize this transition.}
    \label{fig:option}
\end{figure}

Instead of working directly with primitive actions $\a$, one can also work with \emph{temporally extended} actions, commonly referred to as skills, options or templates~\cite{Thrun1994,Sutton1999,Neumann2009}.
While a primitive action is only \sq{active} for a single timestep, bringing the agent from one \emph{state} to the next; options can last multiple timesteps, bringing the agent from one \emph{region} in state space to another.
Once an option is activated, and until it is terminated, the option policy takes over control and provides the environment with the necessary primitive actions at every timestep to complete this transition.
\figref{fig:option} visualizes this process.

\subsubsection*{Definition}
More formally an option $\o$ can be defined by the triplet $\tuple{\I_{\o}, \pi_{\o}, \chi_{\o}}$ with initiation set $\I_{\o} \subset \S$, option policy $\pi_{\o}(\a[0] \given \s[0])$ and termination condition $\chi_{\o}(\s[1])$. The initiation set determines for which regions in state space the option can be activated. Once the option is active, the agent samples actions from the option policy $\a[0] \sim \pi_{\o}(\cdot \given \s[0])$ at every timestep until its termination. In general, the option's termination conditions $\chi_{\o}$ are probabilistic. More precisely, for every state $\s[1]$ the option terminates with probability $\chi_{\o}(\s[1])$ or remains active with probability $1-\chi_{\o}(\s[1])$. In this work we specifically focus on \emph{deterministic} options, which have both a deterministic policy $\a[0] = \pi_{\o}(\s[0])$ and deterministic termination conditions $\chi_{\o}(\s[1]) \in \set{0,1}$. Such deterministic termination conditions effectively partition the state space, allowing us to define the termination set $\T_{\o} = \set{\s[1] \in \S \given \chi_{\o}(\s[1]) = 1}$.

Remark that primitive actions $\a$ can be seen as a special case under this options framework.
In particular, the option $\tuple{\S, \ind_{\set{\a}}, 1}$ is effectively equivalent to the primitive action $\a$, as it can be selected for all states $\I = \S$, has deterministic policy $\pi(\s[0])=\a$ for all states $\s[0]$, and terminates after one timestep as $\chi(\s[1])=1$ for all states $\s[1]$ ($\T = \S$).

\subsubsection*{Hierarchical Control}
Providing the agent with a set of options $\O$ to choose from, rather than the set of actions $\A$, naturally leads to a hierarchical control setup, as shown in \figref{fig:hierarchical_control} and explained in \ssecref{ssec:hco}.
At the lowest level, the policy of the active option $\o[0] \in \O$ is now responsible for taking the appropriate actions at every timestep $t$, similar to the policy over actions in a classical setup.
At the highest level, a novel \emph{policy over options} or \emph{master policy} $\pol{m}(\o[0] \given \s[0])$ is introduced, describing the probability of activating option $\o[0]$ in state $\s[0]$.
This master policy can only select from the set of available options for a particular state, denoted by $\Oav(\s) = \set{\o \in \O \given \s \in \I_{\o}}$, which is fully determined by each of the option's initiation sets.

Remark that the policies at each level operate on different timescales.
The active option's policy $\pi_{\o[0]}$ operates on the finest timescale, selecting a new action at every timestep; whereas the policy over options $\pol{m}$ operates on a coarser timescale, only selecting a new option on termination of the previously active option.
More precisely, $\o[1] \sim \pol{m}(\cdot \given \s[1])$ when $\o[0]$ terminates in $\s[1]$, and $\o[1] = \o[0]$ otherwise.
From the higher level's perspective, we are thus operating in a \emph{Semi-MDP}~\cite{Puterman1994}, with the options providing the necessary link back and forth to the underlying MDP.

\subsubsection*{State-Option Value Function}
Analogous to the state-action value functions for policies over actions, we can define the \emph{state-option} value function for policies over options $q^{\pol{m}}(\s, \o) = \expect_{\O, \pol{m}, \tau}{\sum_{k=0}^{\infty} \gamma^k r(\s[k], \a[k], \s[k+1]) \given \s[0]=\s,\o[0]=\o}$.
In this case, the expectation is taken over the probability distributions describing the active options $\o[1] \sim (1 - \chi_{\o[0]}(\s[1])) \ind_{\set{\o[0]}}(\cdot) + \chi_{\o[0]}(\s[1])\pol{m}(\cdot \given \s[1])$, the selected actions $\a[0] \sim \pi_{\o[0]}(\cdot \given \s[0])$, and the state transitions $\s[1] \sim \tau(\cdot \given~ \s[0], \a[0])$ for all timesteps.

\AuthorCite{Sutton1999} describe various strategies for finding the optimal master policy $\pol{m}^*$. The simplest approach is \emph{option-to-option} value learning, which is based on the optimal Bellman equation for the superimposed Semi-MDP, $q^*(\s, \o) = \expect_{\pi_{\o}, \tau}[\big]{\sum_{k=0}^{d-1} \gamma^k r(\s[k], \a[k], \s[k+1]) + \gamma^d \max_{\o[d] \in \Oav(\s[d])} q^*(\s[d], \o[d]) \given \s[0]=\s, \o[0]=\o}$, with $d$ the duration of $\o$ from $\s$. In other words, this method considers the execution of an option, from activation until termination, as an indivisible operation. In this work we specifically focus on the alternative \emph{intra-option} value learning technique, which \sq{breaks up} the options in the finer timescale of the underlying MDP and also provides value updates for the intermediately visited states throughout an option's execution. When working with deterministic options, the optimal value function can then be written down as $q^*(\s, \o) = \expect_{\tau}{r(\s[0], \a[0], \s[1]) + \gamma u^*(\s[1], \o[0]) \given \s[0]=\s, \o[0]=\o, \a[0]=\pi_{\o}(\s[0])}$ with
\begin{equation*}
    u^*(\s[1], \o[0]) = \begin{dcases}
    q^*(\s[1], \o[0]) & \s[1] \notin \T_{\o[0]} \\
    \max_{\o[1] \in \Oav(\s[1])} q^*(\s[1], \o[1]) & \s[1] \in \T_{\o[0]}
    \end{dcases}.
\end{equation*}
In \secref{sec:methods} we show how a practical estimate of this optimal value function and the corresponding optimal master policy can be found using deep constrained Q-learning~\cite{Kalweit2020}. Following the intra-option technique instead of the option-to-option approach will turn out to be pivotal for supporting combined and hybrid options.

\subsection{Methods}
\begin{figure*}
    \centering
    \includegraphics[width=0.8\textwidth]{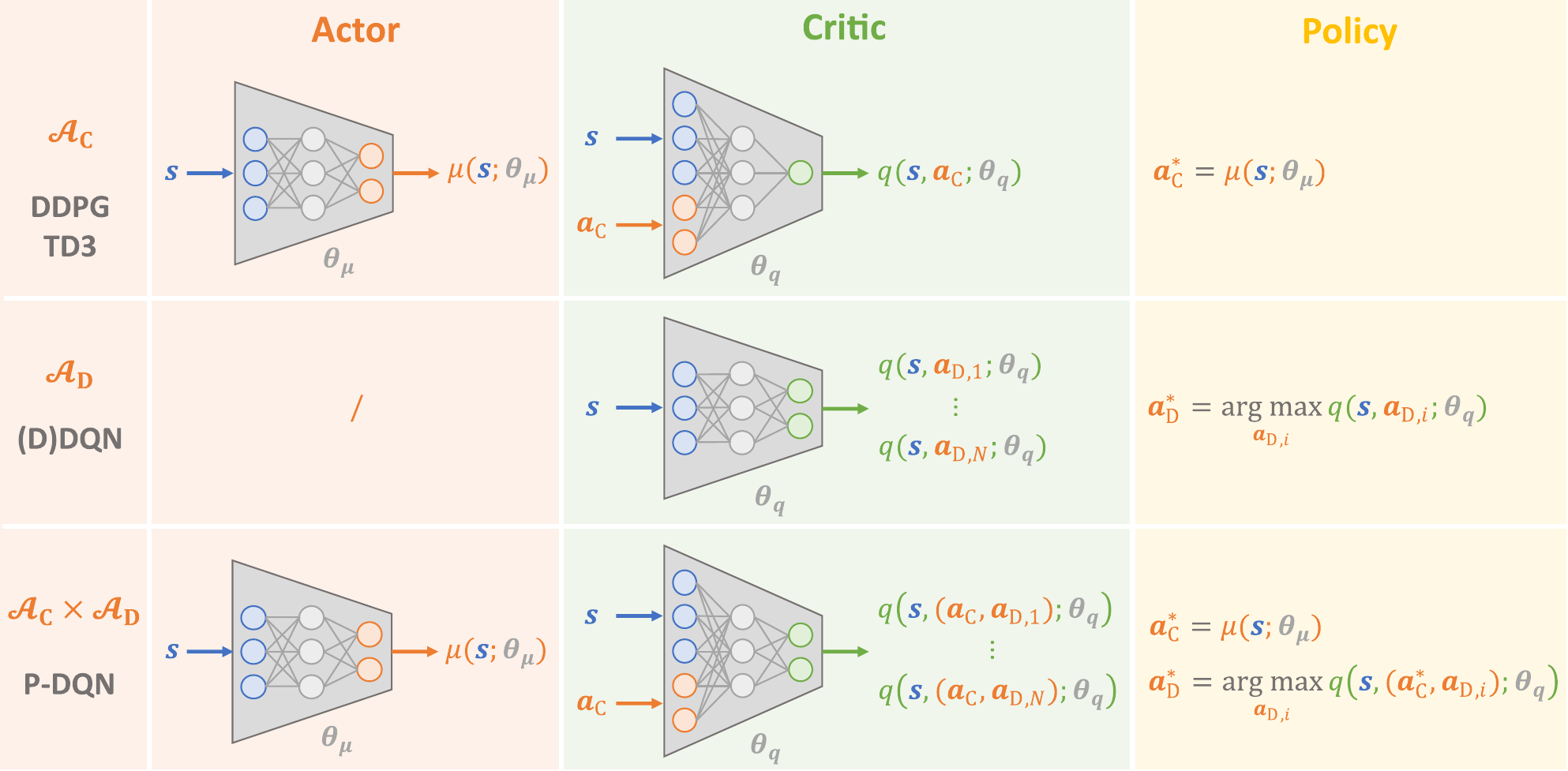}
    \caption{Overview of the model architectures and derived policies for the methods discussed in this paper. Top: actor-critic methods for policies over continuous actions $\a{c} \in \A{c}$. Middle: value-based methods for policies over discrete actions $\a{d} \in \A{d}$. Bottom: actor-critic methods for policies over hybrid actions $\tuple{\a{c}, \a{d}} \in \A{c} \times \A{d}$.}
    \label{fig:methods}
\end{figure*}

There are many ways to find an approximate optimal policy for \eqref{eq:rl_objective}.
In this paper, we specifically limit ourselves to some recent model-free methods for learning deterministic policies.
In particular, due to the discrete nature of option selection, we primarily focus on methods that support discrete ($\A = \A{d} \subseteq \O$) or hybrid ($\A = \A{c} \times \A{d}$) action spaces. Policies over continuous action spaces ($\A = \A{c}$) are only considered as a baseline to compare against in the conducted experiments.
\figref{fig:methods} provides an overview of the different setups used throughout this work.

\subsubsection*{Actors and Critics}
Two categories of methods can be immediately distinguished from this figure. On the one hand, \emph{actor-critic} methods jointly train two deep neural networks: an actor $\mu(\s; \p_\mu)$ and a critic $q(\s, \a; \p_q)$. The actor network represents a deterministic policy over \emph{continuous} actions, $\a{c} = \pol{c}(\s) = \mu(\s; \p_\mu)$, approximating the optimal policy; whereas the critic network approximates the optimal value function.
On the other hand, \emph{value-based} methods only train one deep neural network, the Q-Network (critic) $q(\s, \a; \p_q)$, which approximates the optimal value function. The optimal policy over \emph{discrete} actions is not explicitly learned, but rather derived as the greedy policy for this learned value function, i.e. $\a{d} = \pol{d}(\s) = \argmax_{\a} q(\s, \a; \p_q)$.

The critic network is typically trained by minimizing a squared temporal difference error, resulting in the critic loss
\begin{equation}
    L_q(\p_q) = \expect\nolimits_{\exptup \sim \buf}\bracks[\big]{(q(\s[0], \a[0]; \p_q) - y_t)^2},
    \label{eq:critic_loss}
\end{equation}
where the target value $y_t$ can be calculated in various ways, depending on the specific method.
To improve the training stability, these target values are kept steady throughout training through the use of extra target networks. The parameters of these target networks are denoted with a prime, and are updated more slowly using Polyak averaging $\p_i' \leftarrow \tau \p_i + (1 - \tau) \p_i'$.
The actor network is trained by maximizing an expected discounted return estimate, usually involving an evaluation of the critic network. The specific definition of the actor loss $L_\mu(\p_\mu)$ depends on the method.

Although there are quite some similarities between the various approaches, there is an important difference in the architecture of the critic model, as illustrated in \figref{fig:methods}.
The continuous actions in the actor-critic setup are passed as an input to the critic model, which has a single output neuron corresponding to the value for the given state-action pair.
On the other hand, in the value-based setup, only the state is passed as an input to the critic model, which has a dedicated output neuron for each discrete action to denote the value for the given state and respective action.
When working with hybrid action spaces, a mixed critic architecture is used, which takes the continuous part of the action as an input (together with the state), and has separate output neurons for the discrete part of the action.
Further details about each of the used methods are discussed in the remainder of this section.

\subsubsection*{Continuous Actions}
For continuous action spaces, we work with actor-critic methods, such as \sq{Deep Deterministic Policy Gradient} (DDPG)~\cite{Lillicrap2016} and \sq{Twin Delayed DDPG} (TD3)~\cite{Fujimoto2018}.
To ensure sufficient exploration of the state-action space, the Gaussian behavioural policy $\gauss(\pi(\s), \sigma^2 I)$ is deployed to populate the replay buffer.
The critic network is trained by minimizing the critic loss \eqref{eq:critic_loss} with target value $y_t = r_t + \gamma q(\s[1],\mu(\s[1];\p_\mu');\p_q')$ for the DDPG method.
The actor network is updated simultaneously by minimizing the actor loss $L_\mu(\p_\mu) = -\expect_{\s[0] \sim \buf}{q(\s[0],\mu(\s[0];\p_\mu);\allowbreak\p_q)}$.

\paragraph*{TD3}
The performance and training stability of the original DDPG method can be further improved according to \AuthorCite{Fujimoto2018}.
In their proposed TD3 algorithm, extra twin networks are introduced, actor and target networks are updated less frequently than critic networks, and smoothed (regularized) target values are used.
More precisely, two separate critic networks (the twins), denoted by parameters $\p_{q,1}$ and $\p_{q,2}$, are jointly trained using the same critic loss \eqref{eq:critic_loss} but with altered target values $y_t = r_t + \gamma \min_{i \in \set{1, 2}} q(\s[1], \mu(\s[1]; \p_\mu') + \epsilon_\mathrm{c}; \p_{q,i}')$ with $\epsilon_\mathrm{c}$ a noise sample from the clipped Gaussian distribution $\gauss(\bm{0}, \sigma_\mathrm{c}^2 I)|_{\civ{-c}{c}}$.

\subsubsection*{Discrete Actions}
For discrete action spaces, we resort to value-based approaches based on the popular \sq{Deep Q-Network} (DQN)~\cite{Mnih2015}.
To sufficiently explore the state-action space and fill the replay buffer, the $\epsilon$-greedy behavioural policy is deployed.
The model parameters can then be updated using samples from the replay buffer by minimizing the same critic loss \eqref{eq:critic_loss} as before, although with a different calculation of the target values $y_t$.
For the DQN method, the target values are calculated as $y_t = r_t + \gamma \max_{\a[1]} q(\s[1], \a[1]; \p_q')$.

\paragraph*{Clipped Double DQN}
The performance of the original DQN method can be further improved by reducing overestimation bias~\cite{VanHasselt2016, Fujimoto2018}.
To this end, extra twin networks are introduced and the calculation of the target values is altered.
More precisely, two separate Q-Networks (the twins), denoted by the parameters $\p_{q,1}$ and $\p_{q,2}$, are jointly trained.
The policy is derived from the first twin, i.e. $\pi(\s) = \argmax_{\a} q(\s, \a; \p_{q,1})$.
The same loss function \eqref{eq:critic_loss} is used to train each of the twins, but the target value is now calculated as $y_t = r_t + \gamma \min_{i \in \set{1, 2}} q(\s[1], \pi(\s[1]); \p_{q,i}')$.

\paragraph*{Constrained DQN}
Another DQN extension, the \sq{Constrained DQN} method, is proposed by \AuthorCite{Kalweit2020} and aims to properly handle constrained (state-dependent) action spaces $\A(\s) \subset \A$.
This is done by deriving the policy as the \emph{pruned} greedy policy $\pi(\s) = \argmax_{\a \in \A(\s)} q(\s, \a; \p_q)$.
\AuthorCite{Kalweit2020} illustrate that this pruning is also necessary in the calculation of the target values, $y_t = r_t + \gamma \max_{\a[1] \in \A(\s[1])} q(\s[1], \a[1]; \p_q')$, as otherwise suboptimal policies are retrieved.

\subsubsection*{Hybrid Actions}
There are various approaches to deal with hybrid action spaces, depending on the specific application \cite{Masson2016,Fan2019,Neunert2020}. In this work we exclusively train agents in PAMDPs which have the same continuous parameters for each discrete action, i.e. $\A = \A{c} \times \A{d}$. The \sq{Parameterized DQN} (P-DQN)~\cite{Xiong2018} method was chosen to deal with such hybrid action spaces, as its architecture and methodology resemble the previously discussed TD3 and DQN methods the most.

\paragraph*{Parameterized DQN}
In the P-DQN method, the hybrid policy can be split into a discrete and continuous part, $\a = \pi(\s) = \tuple{\pol{c}(\s), \pol{d}(\s)} = \tuple{\a{c}, \a{d}}$.
Similar to the previously discussed actor-critic methods, two deep neural networks are jointly trained: a critic $q(\s, \tuple{\a{c}, \a{d}}; \p_q)$ for approximating the optimal value function and an actor $\mu(\s; \p_\mu)$ for approximating the optimal policy over the continuous parameters.
The policy over continuous parameters is thus explicitly learned through the actor model $\pol{c}(\s) = \mu(\s; \p_\mu)$, whereas the policy over discrete actions is implicitly derived from the learned critic model as $\pol{d}(\s) = \argmax_{\a{d}} q(\s, \tuple{\pol{c}(\s), \a{d}}; \p_q)$.
To sufficiently explore the state-action space, an $\epsilon$-greedy behavioural policy is used for the discrete actions, and a Gaussian behavioural policy for the continuous parameters.
Once again, the same loss \eqref{eq:critic_loss} can be used for optimizing the critic model, with target values calculated as $y_t = r_t + \gamma \max_{\a{d}[1] \in \A{d}} q(\s[1], \tuple{\a{c}[1], \a{d}[1]}; \p_q')$ and $\a{c}[1] = \mu(\s[1]; \p_\mu')$. The actor model is trained by minimizing $L_\mu(\p_\mu) = -\expect_{\s[0] \sim \buf}{\sum_{\a{d} \in \A{d}} q(\s[0], \tuple{\mu(\s[0];\p_\mu), \a{d}};\allowbreak\p_q)}$.

\section{Autonomous Driving Setup}\label{sec:ads}
The hybrid options framework introduced in the next section is applied to an autonomous driving problem on highways. A proprietary highway simulator is used for this purpose. In this section the most relevant components of this simulator and environment setup are briefly introduced.

\subsection{Objective and Reward}
The objective in this environment is to efficiently navigate highway traffic, while adhering to safety constraints (avoid crashes), driving regulations (keep right) and comfort constraints (smooth movements). A three lane highway with both straight and curved segments is considered for all experiments. Various vehicles are randomly distributed along the different available lanes and are assigned different target velocities at initialization. The virtual driver (agent) is in control of a single vehicle, assigned to drive the maximum allowed speed.

The objective and soft constraints are encoded in the reward signal as a weighted sum of different penalties
\begin{equation*}
    r = \frac{\rw{F} \rc{F} + \rw{V} \rc{V} + \rw{C} \rc{C} + \rw{R} \rc{R}}{\rw{F} + \rw{V} + \rw{C} + \rw{R}}.
\end{equation*}
In order of appearance, these components give a penalty for insufficient following distance ($\rc{F}$), inappropriate speed ($\rc{V}$), drifting from lane center ($\rc{C}$) and not keeping right ($\rc{R}$). An extra penalty is given when the virtual driver collides with other vehicles or the highway boundaries.

\subsection{Vehicle Dynamics and Policies}
The kinematic bicycle model (KBM)~\cite{Rajamani2012} is used to update each vehicle's state, consisting of a global $x$ and $y$ position, heading angle $\psi$ and speed $v$,
\begin{equation*}
    \mat{\dot{x} \\ \dot{y} \\ \dot{\psi} \\ \dot{v}} = \mat{v\cos{\left(\psi+\zeta\right)} \\ v\sin{\left(\psi+\zeta\right)} \\ \frac{v}{l_r}\sin{\zeta} \\ \frac{a}{\cos{\zeta}}}
    \quad
    \zeta = \arctan\left(\frac{l_r}{l_f+l_r}\tan{\delta}\right).
\end{equation*}
The inputs of this dynamical system are the steering angle $\delta$ and the longitudinal acceleration $a$.
In this work, we do not further consider such \emph{global} state vectors and low-level inputs; instead \emph{projected} state vectors (into the road's frame of reference) and high-level actions are used.
Henceforth, we denote the ego vehicle's projected longitudinal velocity (along the road) by $v$, and its projected lateral offset (with respect to the road edge) by $d$.
Dedicated vehicle motion controllers aid in stabilizing the vehicle on the road and facilitate the driving task of the virtual driver. These controllers track the references (actions) $\a$, consisting of a desired relative longitudinal velocity $\Delta v$ and relative lateral position $\Delta d$, set by the driving policies.

To take correct high level steering decisions, the virtual driver needs some extra information about other traffic participants in its neighbourhood. This information is gathered in the agent's observation vector $\s$, containing local information such as the vehicle's velocity components and its offset with respect to different lane centers, and relative information such as relative distances and velocities with respect to neighbouring traffic.
The relative offsets with respect to the lane center of the current lane ($\Delta L = 0$) and nearest lanes on the left ($\Delta L = 1$) and right ($\Delta L = -1$) side of the ego vehicle are referred to as $c_{\Delta L}^{}$. In case there is no subsequent lane on a particular side, $c_{\pm 1}^{}$ is set equal to $c_0^{}$.

Every vehicle is controlled by a policy, mapping observations $\s$ to suitable actions $\a$. The policy of the autonomous vehicle is learned using any of the described reinforcement learning methods in this paper. The policies of the other vehicles in the simulation environment are fixed beforehand. A mixture of vehicles equipped with a custom rule-based policy and a policy implementing the \sq{Intelligent Driver Model} (IDM)~\cite{Treiber2000} and \sq{Minimizing Overall Braking Induced by Lane change} (MOBIL)~\cite{Kesting2007} is used. Both policies try to mimic rudimentary human driving behaviour, although being fully deterministic.

\subsection{Safety and Comfort constraints}
The safety and comfort requirements are not encoded in the reward signal as soft constraints, but dealt with separately as hard constraints. For policies over continuous actions (directly providing setpoints for the motion controllers), safety is ensured through state-dependent action bounds as in De Cooman et al. \cite{DeCooman2023}. These action bounds $\lb{\a}(\s) \le \a \le \ub{\a}(\s)$ are derived from the braking criterion
\begin{equation}
    \min\left(\Delta x, \Delta x + \frac{v_\mathrm{L}^2}{2b} - \frac{v_\mathrm{F}^2}{2b}\right) > \Delta x_\mathrm{SAFE}^{},
    \label{eq:braking_crit}
\end{equation}
where $\Delta x$ is the following distance between $2$ vehicles, $v_\mathrm{L}^{}$ and $v_\mathrm{F}^{}$ are the velocities of the leading and following vehicle respectively, $b$ is the maximum deceleration of both vehicles, and $\Delta x_\mathrm{SAFE}^{}$ is a minimum safe distance to keep between both vehicles. Complying with this braking criterion ensures the following vehicle is always able to avoid a collision with the leading vehicle, under some reasonable worst-case assumptions \cite{DeCooman2023}.
To indicate that an action $a$ is mapped to the safe interval defined by the action bounds, the following (clipping) notation is used $a|_{\civ{\lb{a}}{\ub{a}}} = \min\bracks{\max\parens{a, \lb{a}}, \ub{a}}$.

In this work, we employ the same braking criterion to enforce safety on policies using (hybrid) options. Moreover, under this options framework, additional comfort constraints (smooth lane changes) can be readily embedded in dedicated manoeuvre policies, as discussed in the next section.

\section{Methodology}\label{sec:methods}
To restrict the driving behaviour of the virtual driver, we specifically predefine a set of options for longitudinal and lateral control of autonomous vehicles. Each option's policy automatically provides the necessary actions to perform certain manoeuvres over the course of multiple timesteps, while adhering to the given hard constraints by design. For example, safety of the learned driving policies can be ensured at the finest (MDP) timescale, by reusing the state-dependent action bounds for policies over continuous actions. The virtual driver can thus choose from a comprehensive collection of safe and comfortable manoeuvres (options) based on the current traffic situation, instead of selecting individual low-level actions. This makes the options framework especially suitable for autonomous driving applications, and naturally leads to a hierarchical control architecture, as visualized in \figref{fig:hierarchical_control}.

\begin{table*}[t]
	\caption{Overview of the introduced options for autonomous driving.}
	\label{tab:opts}
    \newcommand\inactive[1]{\multicolumn{1}{>{\columncolor[gray]{0.95}[2pt]} l}{\color[gray]{0.5}{#1}}}
    \newcommand\seprule{%
    \addlinespace%
    }
    \begin{centerbox}
	\begin{tabular}{@{\,} l @{\;} l l l l l @{\,}}
		\toprule
		\multicolumn{2}{@{\,} l}{\textbf{Manoeuvre Option}} & \textbf{Target Velocity }$\mathbf{\vt}$ & \textbf{Target Offset }$\mathbf{\dt}$ & \textbf{Initiation Set }$\mathbf{\I}$ & \textbf{Termination Set }$\mathbf{\T}$ \\
		\midrule
        Emergency & $\o{e}$ & $0|_{\civ{v + \Delta\lb{v}}{v + \Delta\ub{v}}}$ & $d|_{\civ{d + \Delta\lb{d}}{d + \Delta\ub{d}}}$ & $\S$ & $\S$ \\
        \seprule
		Maintain & $\o{m}$ & $v$ & $d$ & $\set{\s \in \S \given \mathrm{Safe}\bracks{\s; \vt, \dt}}$ & $\S$ \\
        \seprule
        \makecell[l]{Velocity Change ($-$) \\[2pt] Velocity Change ($+$)} & \makecell[l]{$\o{vd}$ \\[2pt] $\o{vi}$} & \makecell[l]{$\ceil[\big]{\frac{v}{\delta_v} - 1} \delta_v$ \\[2pt] $\floor[\big]{\frac{v}{\delta_v} + 1} \delta_v$} & \inactive{$d$} & $\set{\s \in \S \given \mathrm{Safe}\bracks{\s; \vt, \dt}}$ & $\set*{\s \in \S \given \begin{aligned}\abs{\vt - v} < \epsilon_v^{} \mspace{74mu} \\ {}\lor \lnot \mathrm{Safe}\bracks{\s; \vt, \dt}\end{aligned}}$ \\
        \seprule
		\makecell[l]{Lane Change (left) \\[2pt] Lane Change (right)} & \makecell[l]{$\o{ll}$ \\[2pt] $\o{lr}$} & \inactive{$v$} & \makecell[l]{$d + c_1^\prime$ \\[2pt] $d + c_{-1}^\prime$} & $\set*{\s \in \S \given \begin{aligned}\abs{\dt - d} \ge \epsilon_d^{} \land v \ge v_\mathrm{m}^{} \\ {}\land \mathrm{Safe}\bracks{\s; \vt, \dt}\end{aligned}}$ & $\set*{\s \in \S \given \begin{aligned}\abs{\dt - d} < \epsilon_d^{} \lor v < v_\mathrm{m}^{} \\ {}\lor \lnot \mathrm{Safe}\bracks{\s; \vt, \dt}\end{aligned}}$ \\
		\bottomrule
	\end{tabular}
    \end{centerbox}
\end{table*}

\subsection{Safe Options for Autonomous Driving}\label{ssec:ado}
We start with a concise description of the various predefined options tailored to autonomous driving tasks.
A virtual driver typically needs to consider two high-level control decisions at every moment in time: longitudinal (velocity) control and lateral (position) control. For both of these control tasks, dedicated options are introduced and summarized in \tabref{tab:opts}. The set of options acting on the longitudinal velocity is denoted by $\O{v} = \set{\o{e}, \o{m}, \o{vd}, \o{vi}}$, whereas the set of options acting on the lateral position is denoted by $\O{d} = \set{\o{e}, \o{m}, \o{ll}, \o{lr}}$.
Once activated, the option's policy provides the vehicle motion controllers with the necessary actions to perform a certain predefined manoeuvre, bringing the ego vehicle from one region in state space to another over the course of multiple timesteps (e.g. a lane change manoeuvre). The virtual driver only needs to select another option after the active option terminates, either after successfully completing its associated manoeuvre, or after an abortion caused by other termination conditions (such as safety constraint violations). Although this manoeuvre is limited to either a longitudinal or lateral movement for the options introduced in this work, our methodology allows to readily combine multiple options (and their manoeuvres), as outlined in \ssecref{ssec:hco}.

\subsubsection*{Option Targets}
The effect of each option on the longitudinal velocity or lateral position is indicated by the target velocity $\vt$ and target offset $\dt$. The associated option policy provides the necessary actions to achieve these targets from any initial state, while adhering to hard safety and comfort constraints during the manoeuvre's execution \emph{by design}. For example, the following option policy could be used for this purpose, $\pi_{\o}(\s) = \mat{\vt - v ; \dt - d}|_{\civ{\lb{\a}(\s)}{\ub{\a}(\s)}}$, as it takes safety constraints into account by enforcing the state-dependent action bounds (but disregards any other comfort constraints). Remark that for the Markov property to hold, it is essential to have fixed targets for all states, regardless of whether the option is already active or not in a particular state.

The option targets are also used to assess the safety of the option's associated manoeuvre. More precisely, the manoeuvre is considered to be safe for a certain state if the target velocity and position are within the state-dependent action bounds for that state, i.e. $\mathrm{Safe}\bracks{\s; \vt, \dt} = \Delta\lb{v} \le \vt - v \le \Delta\ub{v} \land \Delta\lb{d} \le \dt - d \le \Delta\ub{d}$. In our (model-free) setup, this effectively corresponds to verifying that the braking criterion \eqref{eq:braking_crit} holds for all interpolated states between the current state and an estimated target state at the end of the manoeuvre. This estimated target state is simply derived from the current state by instantaneously moving the ego vehicle to the target position and immediately adopting the target velocity, while leaving all other state components unchanged. For model-based methods, the learned environment model could be leveraged instead to obtain more accurate state predictions, for which the braking criterion can be checked (or from which action bounds can be derived).

\subsubsection*{Option Safety}
Safety of the various options is then ensured by extending the termination sets and restricting the initiation sets.
More specifically, if the planned manoeuvre is considered unsafe in a certain state, that state is added to the termination set and removed from the initiation set.
As a result, an option can only be activated if it will not bring the virtual driver to an unsafe state during the option's execution (according to the used prediction model); and it is terminated as soon as new predictions, obtained throughout the option's execution, become unsafe.
Remark that the initiation and termination sets do not need to be fully constructed beforehand, rather the initiation and termination conditions can be evaluated for each individual state encountered by the virtual driver. By reevaluating the safety conditions at every timestep, the latest available state information is always used in deciding whether to activate or terminate an option.

\subsubsection*{Option Interruption}
Most of the introduced options run until their targets are reached or their safety conditions are violated. To prevent unlimited execution and to ensure other options can be selected at any time, certain \emph{primitive} options are also defined, which are automatically terminated after every time step ($\T = \S$). An alternative implementation could make the options \emph{interruptable} instead \cite{Sutton1999}, such an extension is however not further investigated in this work.

\subsubsection*{Overview}
The various options from \tabref{tab:opts} are succinctly described below.
Safety and comfort constraints are embedded in each option's policy, as well as through its termination and initiation sets.

\paragraph*{Emergency Option}
To ensure that at least one option is always available, the emergency option $\o{e}$ is introduced.
This option tries to slow down the vehicle as fast as possible ($\vt \rightarrow 0$) and tries to maintain the current lateral position ($\dt \rightarrow d$), while adhering to the safety constraints.
For any encountered unsafe states, the vehicle is moved towards the nearest safe region, by clipping the target values using the state-dependent action bounds.
Although it is the only option that is always available ($\I = \S$), it should be considered as a fallback for emergency situations; as generally, other, more suitable options should be available.
Therefore, it is defined as a primitive option ($\T = \S$), ensuring this emergency manoeuvre can be interrupted at any time.

\paragraph*{Maintain Option}
Additionally, we specify the primitive maintain option $\o{m}$, which maintains the current longitudinal velocity ($\vt = v$) and lateral position ($\dt = d$).
This option is only available when it is safe to keep driving at the current velocity and position ($\mathrm{Safe}\bracks{\s; \vt, \dt}$); and it is terminated after every time step ($\T = \S$), allowing other options to be selected at any moment.

\paragraph*{Longitudinal Options}
For longitudinal control, we define two \emph{velocity change} options, which decrease ($\o{vd}$) or increase ($\o{vi}$) the current velocity with fixed increments\footnotemark\addtocounter{footnote}{-1}\addtocounter{Hfootnote}{-1} $\delta_v = \qty{2}{m/s}$.
These options provide the necessary actions to attain the target velocity $\vt$ in a safe and comfortable manner. They can only be initiated when the target velocity is considered to be safe ($\mathrm{Safe}\bracks{\s; \vt, \dt}$); and they terminate once the target velocity is reached ($\abs{\vt - v} < \epsilon_v^{} = \qty{0.01}{m/s}$) or when it is no longer safe to drive at the target velocity ($\lnot \mathrm{Safe}\bracks{\s; \vt, \dt}$).

\paragraph*{Lateral Options}
For lateral control, we similarly define two \emph{lane change} options, providing the necessary actions to steer the ego vehicle towards the nearest lane center to its left ($\o{ll}$) or right ($\o{lr}$) in approximately $\qty{5}{s}$ (for a complete lane change at maximum velocity).
The relative offsets to the nearest lane center on the left and right of the ego vehicle are denoted by $c_1^\prime$ and $c_{-1}^\prime$.
Both of these are often equal to their non-primed counterparts, unless the vehicle is not properly centered within its current lane and the current lane's center is the nearest in the respective direction (in which case $c_{\pm 1}^\prime = c_0^{}$)\footnote{This is necessary for the Markov property to hold.}.
These options can only be activated when the car is not yet centered in the target lane ($\abs{\dt - d} \ge \epsilon_d^{} = \qty{0.05}{m}$), and the predicted target state is safe ($\mathrm{Safe}\bracks{\s; \vt, \dt}$).
Additionally, a minimum velocity condition ($v \ge v_\mathrm{m}^{} = \qty{3}{m/s}$) is enforced to ensure that the lane change manoeuvre can be completed in a reasonable timeframe, as otherwise the option could remain active indefinitely.
The option terminates once aligned with the target lane's center ($\abs{\dt - d} < \epsilon_d^{}$), or once the minimum velocity is no longer attained ($v < v_\mathrm{m}^{}$), or when the predicted target state is no longer considered to be safe ($\lnot \mathrm{Safe}\bracks{\s; \vt, \dt}$).

\begin{figure*}
    \centering
    \includegraphics[width=\linewidth]{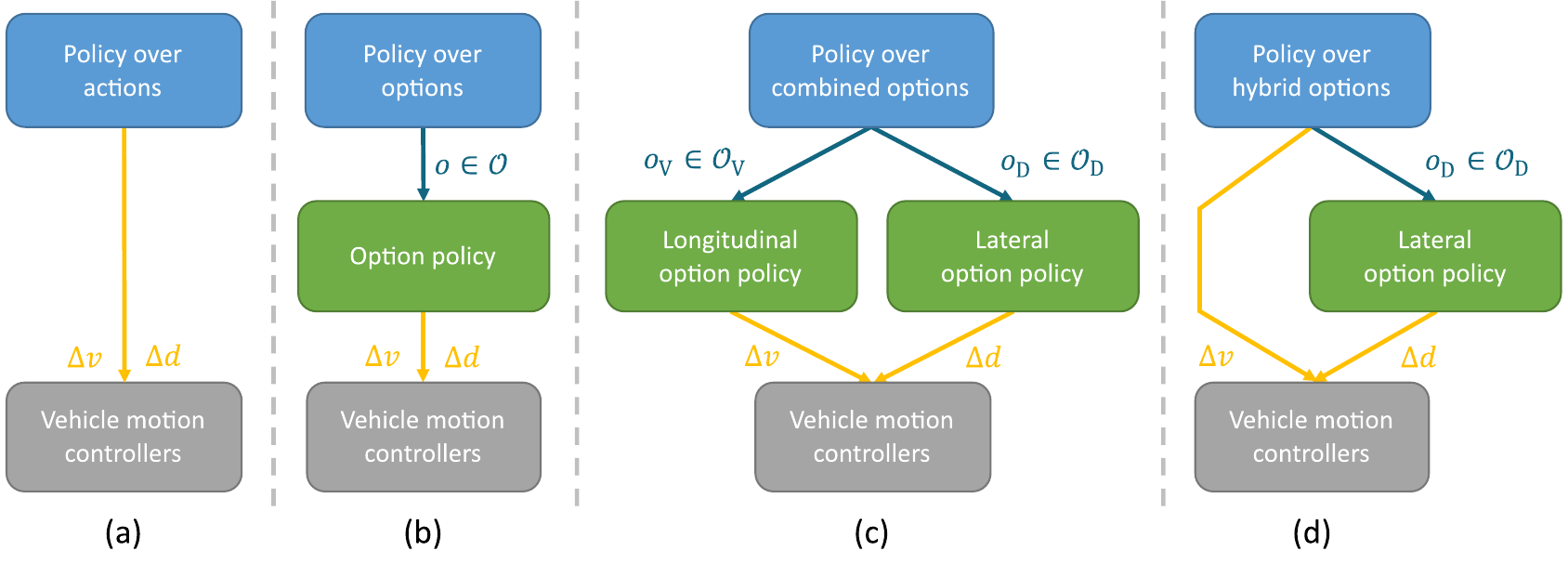}
    \caption{Schematic overview of the different autonomous driving policies used in this work.
    On the left (a) the baseline policy over primitive actions $\tuple{\Delta v, \Delta d}$, directly interfacing with the vehicle motion controllers. On the right (b+c+d) different configurations of policies over options, which select one (or two) new option(s) to execute after a previous one terminated. Once an option policy is in control, it provides the vehicle motion controllers with suitable actions to perform its associated manoeuvre and reach the targets $\vt$ and $\dt$ (see \figref{fig:option} and \tabref{tab:opts}).}
    \label{fig:hierarchical_control}
\end{figure*}

\subsection{Hierarchical Control with Options}\label{ssec:hco}
The previously described options provide the virtual driver with several safe and comfortable manoeuvres for the control of autonomous vehicles. These options can be applied in various hierarchical control architectures. \figref{fig:hierarchical_control} shows an overview of the different setups investigated in this work.

\subsubsection*{Baseline}
The different setups for hierarchical control with options are compared with other approaches using the baseline setup (\figref{fig:hierarchical_control}.a). In this baseline setup, a \sq{classical} policy over primitive actions directly interacts with the vehicle motion controllers through the continuous setpoints $\a = \mat{\Delta v; \Delta d}$.
This policy can either be learned using standard reinforcement learning methods (such as TD3), or be predefined (e.g. following IDM and MOBIL).

\subsubsection*{Options}
With the introduction of dedicated options, the virtual driver obtains the ability to select appropriate manoeuvres to efficiently navigate traffic. In other words, the problem becomes a decision-making task over a discrete set of options (manoeuvres) instead of over a continuous range of actions. We distinguish the higher level actions selected by the master policy, $\a{m} = \pol{m}(\s)$, from the lower level actions that are passed to the vehicle motion controllers, $\a = \pol(\s, \a{m}) = \mat{\Delta v; \Delta d}$.

In the most basic setup (\figref{fig:hierarchical_control}.b), only one option is activated at any time, selected from the complete set of options, $\a{m} = \o \in \O$ ($= \O{v} \union \O{d}$). The selected option takes over control until termination and supplies the vehicle motion controllers with both longitudinal and lateral setpoints, $\mat{\Delta v; \Delta d} = \pi_{\o}(\s)$. The master policy $\pol{m}$ and its behavioural counterpart $\bpol{m}$ only select a new option upon termination of the previously active one (or at the start of the episode).

\subsubsection*{Combined Options}
A drawback of the basic options approach is that only a single manoeuvre can be performed at any given time. For example, the learned policies lack the ability to combine a lateral lane change manoeuvre with a longitudinal acceleration manoeuvre, which can be performed by the continuous baseline policies. To address this limitation, we also propose an extended hierarchical control architecture with support for \emph{combined} options for longitudinal and lateral control (\figref{fig:hierarchical_control}.c). Such a setup can significantly increase the flexibility of the virtual driver, providing it with similar capabilities as the baseline models.
In particular, two options are activated at any time, $\a{m} = \tuple{\o{v}, \o{d}} \in \O{v} \times \O{d}$, one for longitudinal control $\o{v}$ and one for lateral control $\o{d}$. The selected options take over control until termination, and each of them supplies a longitudinal or lateral setpoint to the vehicle motion controllers, $\Delta v = \mat{1 & 0} \pi_{\o{v}}(\s)$ and $\Delta d = \mat{0 & 1} \pi_{\o{d}}(\s)$. As soon as either option terminates, the (behavioural) master policy $\pol{m}$ ($\bpol{m}$) selects a new option for the respective direction of control, while the other non-terminating option remains active.

\subsubsection*{Hybrid Options}
Finally, a hybrid setup (\figref{fig:hierarchical_control}.d) is proposed with the aim of combining the best of the continuous (baseline) and discrete (options) approaches. The resulting setup closely aligns with the decision process of human drivers. More specifically, the hybrid master policy can select a continuous longitudinal velocity parameter, while picking a discrete lateral manoeuvre (option), $\a{m} = \tuple{\Delta v, \o{d}} \in \mathbb{R} \times \O{d}$. The velocity setpoint $\Delta v$ is passed directly to the vehicle motion controllers, while the selected option supplies the lateral position setpoint $\Delta d = \mat{0 & 1} \pi_{\o{d}}(\s)$. Remark that the continuous part of the (behavioural) master policy $\pol{m,c}$ ($\bpol{m,c}$) selects a new velocity setpoint at every timestep, whereas the discrete part of the (behavioural) master policy $\pol{m,d}$ ($\bpol{m,d}$) only selects a new option upon termination of the previously active one.

\subsection{Implementation}\label{ssec:impl}
We apply a combination of existing RL techniques to derive intra-option value learning and actor-critic algorithms for the various hierarchical control setups with options. The generic algorithm is summarized in \algref{alg:generic} and closely follows the training principles of state-of-the-art RL methods. More detailed implementations for each setup are provided in \appref{app:algorithms}[app].
\begin{algorithm}[h!]
    \caption{Hierarchical RL with Options (Generic)}\label{alg:generic}
    \begin{algorithmic}
        \State \textbf{Initialize} critic (and actor) models $\p_{q,i}, \p_{q,i}'$ ($\p_\mu, \p_\mu'$)
        \State $\buf \gets \emptyset$
        \For{each episode}
            \For{each environment step $t$}
                \State $\a{m}[0] \sim \bpol{m}(\cdot \given \s[0], \a{m}[-1])$ \Comment{Select high level action}
                \State $\a[0] = \pi(\s, \a{m}[0])$ \Comment{Obtain low level action}
                \State $\s[1] \sim \tau(\cdot \given \s[0], \a[0])$
                \State $r_t = r(\s[0], \a[0], \s[1])$
                \State $\buf \gets \buf \union \set{\tuple{\s[0], \a{m}[0], r_t, \s[1]}}$
            \EndFor
            \For{each gradient step}
                \State Sample batch $B$ from $\buf$
                \For{all $\tuple{\s[0], \a{m}[0], r_t, \s[1]} \in B$}
                    \State $\a{m}[1] = \pol{m}(\s[1], \a{m}[0])$ \Comment{Calculate targets}
                    \State $y_t = r_t + \gamma \min_{i \in \set{1, 2}} q(\s[1], \a{m}[1]; \p_{q,i}')$
                \EndFor
                \State $\widehat{L}_q(\p_{q,i}; B) = \frac{1}{\abs{B}}\sum_B (q(\s[0], \a{m}[0]; \p_{q,i}) - y_t)^2$
                \State $\widehat{L}_\mu(\p_\mu; B) = \frac{-1}{\abs{B}}\sum\limits_B \sum\limits_{\o{d} \in \Oav{d}(\s[0])} \mspace{-24mu}q(\s[0], \tuple{\mu(\s[0]; \p_\mu), \o{d}}; \p_{q,1})$
                \State $\p_{q,i} \gets \p_{q,i} - \eta_q \nabla_{\p_{q,i}} \widehat{L}_q(\p_{q,i}; B)$ \Comment{Update critics}
                \State $\p_\mu \gets \p_\mu - \eta_\mu \nabla_{\p_\mu} \widehat{L}_\mu(\p_\mu; B)$ \Comment{Update actor}
                \State $\p_m' \gets \tau \p_m + (1 - \tau) \p_m'$ \Comment{Update target models}
            \EndFor
        \EndFor
    \end{algorithmic}
\end{algorithm}

\begin{remark}
    Although in \algref{alg:generic} a sample from the semi-Markov (behavioural) master policy is taken for every timestep, this is often not needed in practice. In particular, the discrete component of the policy should only select a new option upon termination of the active option.
\end{remark}

\subsubsection*{Option Availability}
At several points during training and deployment of the master policy over options, we need to know which options are available to the agent. The set $\Oav(\s)$ provides this information at the time of selecting a new option, but that is insufficient. Since the options investigated in this work are uninterruptible, the only available option is often the currently active one. Therefore, we also define the closely related set of available options for a particular state $\s'$, \emph{given that option $\o$ is active} upon reaching this state. This extended availability set additionally depends on the option's termination sets, and is defined as
\begin{equation*}
    \Oav(\s', \o) = \begin{cases}
        \Oav(\s') & \text{if } \s' \in \T_{\o} \\
        \set{\o} & \text{otherwise}
    \end{cases}.
\end{equation*}

Similarly as for the initiation and termination sets, the availability sets do not need to be fully constructed beforehand, but can be evaluated \sq{on the go} for the encountered states and options. For very complex initiation or termination set conditions, the option availabilities for states $\s[0]$ and $\s[1]$ can also be stored in the replay buffer to speed up the computations (at the cost of increased memory usage).

\subsubsection*{Intra-Option Value Learning}
The critic networks are trained using a variant of the clipped DDQN method~\cite{Fujimoto2018}, adapted for use with options by following the intra-option value learning technique. Additionally, the methodology of the constrained DQN method~\cite{Kalweit2020} is applied to enforce the constrained initiation and termination of the options. Ultimately, both critic networks (twins) can be trained by minimizing the critic loss \eqref{eq:critic_loss} with target values $y_t = r_t + \gamma \min_{i \in \set{1, 2}} q(\s[1], \a{m}[1]; \p_{q,i}')$ and $\a{m}[1]$ obtained from the (constrained) master policy $\pol{m}$.

By breaking up the options into their constituent actions for every timestep, the critic models can be updated from every individual interaction with the environment stored in the replay buffer (similar to classical RL approaches). Moreover, this allows to flexibly combine an action from the option's policy with an action from other policies (either over options or primitive actions) in every timestep, which is not possible in the option-to-option approach.
\begin{remark}
    Even though the option policies have to satisfy the Markov property in order to be able to use the intra-option learning rule, experiments suggested it also works well for option policies that do not always satisfy the Markov property.
\end{remark}

\subsubsection*{Options}
The master policy over single options (\figref{fig:hierarchical_control}.b) is derived from the first critic twin as $\a{m}' = \pol{m}(\s', \a{m}) = \argmax_{\o' \in \Oav(\s', \o)} q(\s', \o'; \p_{q,1})$, or equivalently
\begin{align*}
    \o[1] &= \begin{cases}
        \argmax_{\o' \in \Oav(\s[1])} q(\s[1], \o'; \p_{q,1}) & \s[1] \in \T_{\o[0]} \\
        \o[0] & \text{otherwise}
    \end{cases}.
\end{align*}
The behavioural master policy $\bpol{m}$ is the $\epsilon$-greedy policy derived from $\pol{m}$. Remark that the random selection must take into account the option availabilities, i.e. $\bpol{m}(\a{m}' \given \s', \a{m}) = \epsilon \abs{\Oav(\s', \o)}^{-1}\ind_{\Oav(\s', \o)}(\a{m}') + (1 - \epsilon) \ind_{\set{\pol{m}(\s', \a{m})}}(\a{m}')$.

\subsubsection*{Combined Options}
When working with combined options (\figref{fig:hierarchical_control}.c), the high level action $\a{m}$ is a two-dimensional discrete decision variable. In this case, the master policy can be derived from the first critic twin as ${\a{m}' = \pol{m}(\s', \a{m})} = \argmax_{\a{m}' \in \Oav{v}(\s', \o{v}) \times \Oav{d}(\s', \o{d})} q(\s', \a{m}'; \p_{q,1})$. This policy only selects a new option upon termination of any previously active one, which becomes more clear when rewriting it as
\begin{align*}
    \a{m}' \!=\! \begin{cases}
        \mspace{-6mu}\argmax\limits_{\o{v}' \in \Oav{v}(\s'), \o{d}' \in \Oav{d}(\s')} q(\s, \tuple{\o{v}', \o{d}'}; \p_{q,1}) & \s' \!\in\! \T_{\o{v}} \!\intersec\! \T_{\o{d}} \\
        \mspace{-6mu}\tuple{\argmax_{\o{v}' \in \Oav{v}(\s')} q(\s', \tuple{\o{v}', \o{d}}; \p_{q,1}), \o{d}} & \s' \!\in\! \T_{\o{v}} \!\setminus\! \T_{\o{d}} \\
        \mspace{-6mu}\tuple{\o{v}, \argmax_{\o{d}' \in \Oav{d}(\s')} q(\s', \tuple{\o{v}, \o{d}'}; \p_{q,1})} & \s' \!\in\! \T_{\o{d}} \!\setminus\! \T_{\o{v}} \\
        \mspace{-6mu}\tuple{\o{v}, \o{d}} & \text{otherwise}
    \end{cases}.
\end{align*}
The behavioural master policy $\bpol{m}$ is the $\epsilon$-greedy policy derived from $\pol{m}$, taking into account the option availabilities for the random selection, i.e. $\bpol{m}(\a{m}' \given \s', \a{m}) = \epsilon \abs{\Oav{v}(\s', \o{v}) \times \Oav{d}(\s', \o{d})}^{-1}\ind_{\Oav{v}(\s', \o{v}) \times \Oav{d}(\s', \o{d})}(\a{m}') + (1 - \epsilon) \ind_{\set{\pol{m}(\s', \a{m})}}(\a{m}')$.

\subsubsection*{Hybrid Options}
The implementation for hybrid options (\figref{fig:hierarchical_control}.d) is additionally based on the P-DQN method~\cite{Xiong2018} and has a special (hybrid) structure for the critic networks (see \figref{fig:methods}). Similar to classical TD3 (used in the baseline setup), the calculation of target values $y_t$ is slightly altered to perform target policy smoothing~\cite{Fujimoto2018}. Furthermore, the state-dependent action bounds methodology~\cite{DeCooman2023} is applied to ensure the selected continuous (velocity) parameters are safe.

In this setup, there is an extra actor model for selecting the continuous parameters, explicitly modeling the continuous part of the master policy. More precisely, the actor network provides the normalized parameter $\Delta \tilde{v} = \mu(\s; \p_\mu) \in \civ{-1}{1}$, which is rescaled to $\Delta v = \pol{m,c}(\s) = \sigma_\mathrm{pwl}(\Delta \tilde{v}; \s)$ using the piecewise linear rescaling function with state-dependent action bounds $\civ{\lb{\Delta v}(\s)}{\ub{\Delta v}(\s)}$~\cite{DeCooman2023}.
This actor model is trained by minimizing the actor loss $L_\mu(\p_\mu) = -\expect_{\s[0] \sim \buf}{\sum_{\o{d} \in \Oav{d}(\s[0])} q(\s[0], \tuple{\mu(\s[0];\p_\mu), \o{d}}; \p_{q,1})}$.
The discrete part of the master policy is implicitly derived from the first critic model as $\o{d}' = \pol{m,d}(\s', \o{d}) = \argmax_{\o{d}' \in \Oav{d}(\s', \o{d})}\! q(\s', \tuple{\mu(\s'; \p_\mu),\! \o{d}'}; \p_{q,1})$, or equivalently
\begin{align*}
    \o{d}[1] \!=\! \begin{cases}
        \!\argmax\limits_{\o{d}' \in \Oav{d}(\s[1])}\mspace{-6mu} q(\s[1], \tuple{\Delta \tilde{v}_{t+1}, \o{d}'}; \p_{q,1}) & \s[1] \!\in \T_{\o{d}[0]} \\
        \!\o{d}[0] & \text{otherwise}
    \end{cases}.
\end{align*}

The behavioural master policy $\bpol{m}$ can similarly be split in a continuous and discrete part. For the continuous part, the truncated Gaussian behavioural policy $\gauss_{-1}^{1}(\mu(\s; \p_\mu), \sigma_\epsilon^2)$ is used. For the discrete part, the $\epsilon$-greedy policy derived from $\pol{m,d}$ is used, taking into account each of the option availabilities for the random selection, i.e. $\bpol{m,d}(\o{d}' \given \s', \o{d}) = \epsilon \abs{\Oav{d}(\s', \o{d})}^{-1}\ind_{\Oav{d}(\s', \o{d})}(\o{d}') + (1 - \epsilon) \ind_{\set{\pol{m,d}(\s', \o{d})}}(\o{d}')$.

\section{Experiments}\label{sec:experiments}
All presented methods for hierarchical control with options are compared with baseline policies in the simulated autonomous driving environment described in \secref{sec:ads}. More specifically, $4$ different RL policies are trained: one baseline policy over continuous actions, and three master policies over single, combined, and hybrid options (see \figref{fig:hierarchical_control}). The baseline policy over continuous actions is trained using STD3~\cite{DeCooman2021} (for improved smoothness) with enforced state-dependent action bounds~\cite{DeCooman2023} (for safety). To make the comparisons fair, both the smoothness regularization and action bounds are also applied to the continuous component of the hybrid master policy, as illustrated in \algref{alg:hybrid}[app]. An overview of the used hyperparameters can be found in \appref{app:hyper}[app].

For every configuration, $R=10$ policies are trained, each using a different seed for initialization of the training process. Throughout training, $E=10$ independent evaluation episodes are executed to obtain an estimate of the learned policy's average performance. Out of all $R$ repeats, the best performing policy is selected. This selection is based on the average evaluation reward and only considers policies obtained after $2/3$ of the training process has completed, which guarantees a good performance-smoothness trade-off for the policies over continuous actions. This best policy is separately evaluated on some additional benchmark scenarios for autonomous driving and extensively tested under varying traffic conditions.

\subsection{Training and Evaluation}
\begin{figure}
    \centering
    \includegraphics[width=0.8\columnwidth]{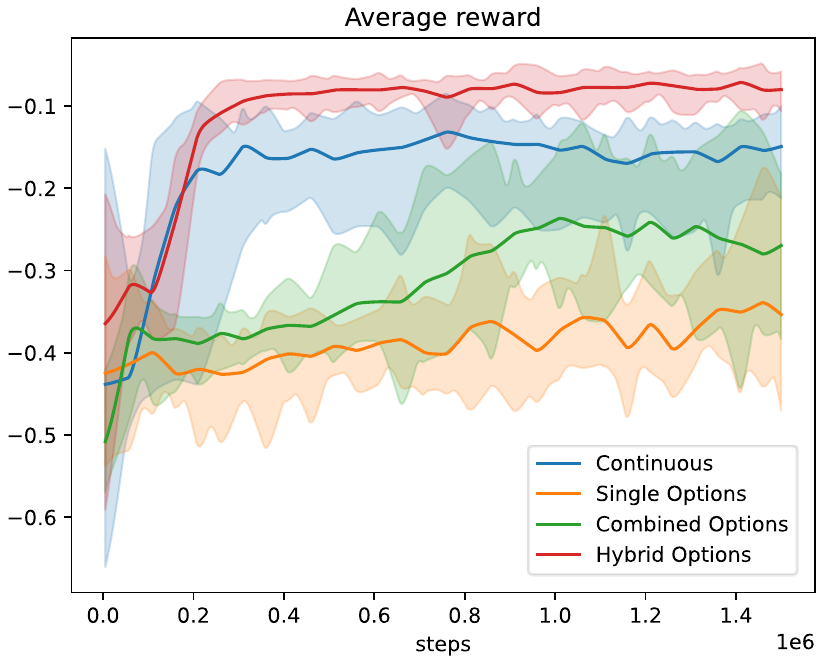}
    \caption{Summarized evaluation performance for each policy throughout training. Solid lines represent the mean value over $10$ repeated experiments, the shaded areas denote the minimum and maximum values.}
    \label{fig:eval}
\end{figure}

The average rewards for each of the $RE$ evaluation episodes and for every configuration are aggregated and plotted (using exponential average smoothing with $\alpha=0.1$) in \figref{fig:eval}. It is clear the hybrid options approach outperforms the other policies, as it has both a higher average reward and lower variance between the different repeated experiments. Both the continuous and hybrid policies quickly converge, after approximately \num{2e5} training timesteps, whereas the single options approach converges much more slowly and might not have fully converged yet at the end of training. Looking at the wall clock time (using the same hardware for training), we observe the fastest training time for the continuous policies. Our implementations of the hierarchical approaches seem to be roughly $10\%$ (single), $20\%$ (combined) and $30\%$ (hybrid) slower on average, although further optimizations could potentially accelerate training. \appref{app:results}[app] additionally visualizes the activity of options during evaluation, which underlines the increased interpretability of the option methods. Finally, it should be stressed that all trained policies are completely safe and cause no crashes (in the used simulation environment) both during training and evaluation.

\subsection{Benchmark Results}
\begin{figure}[t]
    \centering
    \includegraphics[width=0.8\columnwidth]{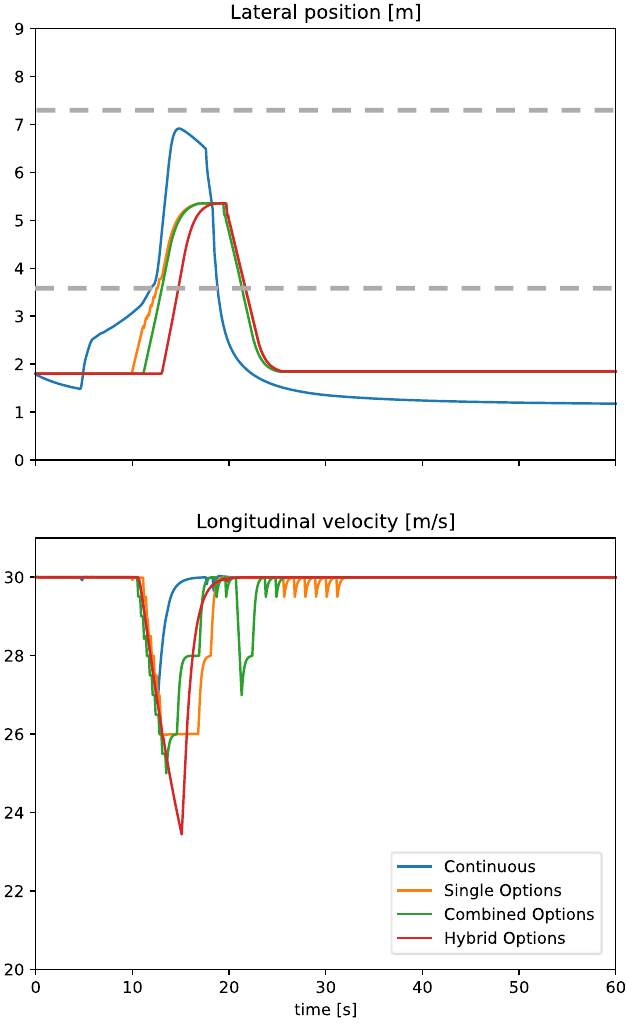}
    \caption{Lateral position and longitudinal velocity plots for each of the selected policies applied to the overtaking benchmark.}
    \label{fig:overtaking}
\end{figure}

The best performing policies for each configuration are separately evaluated on some typical driving tasks and scenarios. \figref{fig:overtaking} shows the observed driving behaviour for the simple scenario of overtaking one slower vehicle in front of the ego vehicle. From the lateral position plot, we can immediately see the benefit of the hierarchical policies over options. By using constrained option policies for lateral control, we obtain better alignment within the lane and much more desirable lane change manoeuvres, with more gradual transitions and without any overshoot. Although the smoothness regularization for the baseline policy over continuous actions already improves passenger comfort to some extent --- by avoiding abrupt and oscillatory lateral movements --- it is still insufficient to obtain the same kind of comfortable and consistent lane change manoeuvres as for the option approaches.

Looking carefully at the velocity plot, some interesting properties of the various option setups are revealed. The velocity profiles for the policies over single and combined options are less smooth and contain some \sq{discrete jumps}, reducing passenger comfort. This phenomenon is caused by the particular definition of the options for longitudinal control, which modify the velocity in fixed increments $\delta_v$ to adhere to the Markov property. Some additional experiments with non-Markov options for longitudinal control (with more flexible velocity adaptation to avoid this inherent drawback), yielded promising initial results.
Slightly harder to spot is the different operation of policies over single options compared to policies over combined options. The former policies can only perform one longitudinal \emph{or} lateral manoeuvre at once, resulting in a constant velocity profile during the lane changes. The latter policies can combine both a longitudinal \emph{and} lateral manoeuvre at the same time, resulting in slightly faster and more efficient movements on the road.
Overall the policy over hybrid options scores the best, with the smoothest velocity and position profiles on this overtaking benchmark.

\subsection{Test Results}
\begin{figure}[t]
    \centering
    \includegraphics[width=0.8\columnwidth]{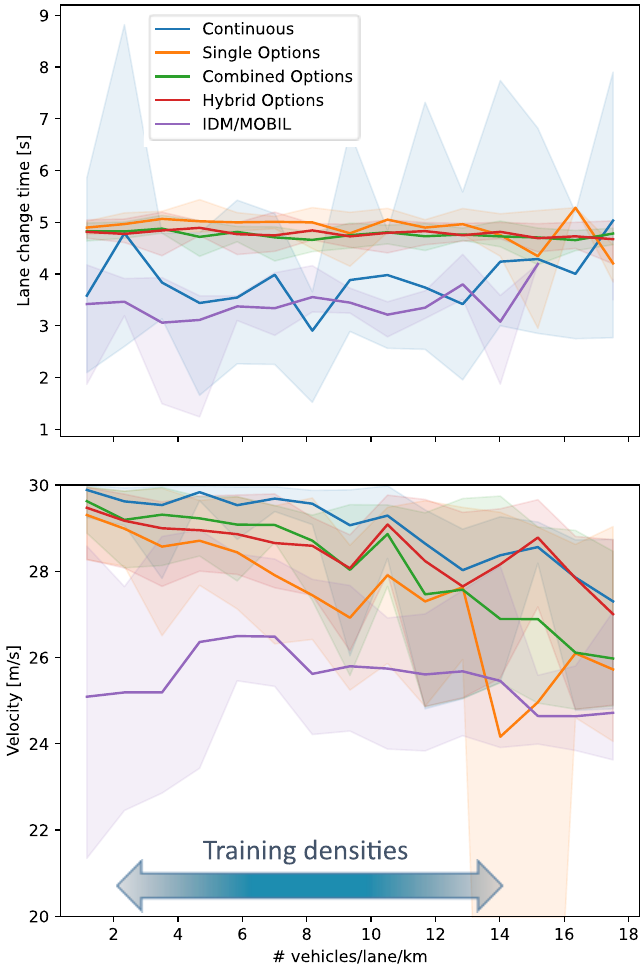}
    \caption{Evaluation of the selected policies in various traffic conditions and comparison with a conservative IDM/MOBIL policy. Solid lines represent the mean value over $10$ episodes, the shaded areas denote the minimum and maximum values.}
    \label{fig:test_results}
\end{figure}

Finally, the best performing policies for each configuration are also extensively tested under various traffic conditions, ranging from very calm to dense traffic scenarios, and compared with a conservative baseline policy following IDM and MOBIL. These test results are summarized in \figref{fig:test_results}, with additional plots provided in \appref{app:results}[app].

Overall, all policies have very comparable performances and outperform the IDM/MOBIL policy in all circumstances. The baseline policy over continuous actions and policy over hybrid options seem to outperform the other policies at higher traffic densities (more congestion), not seen during training. As expected, all hierarchical approaches with constrained options perform more reliable and consistent lane change manoeuvres. In particular, the lane change time is almost constant and takes around $\qty{5}{s}$, complying with the intended comfort constraints. In contrast, the baseline policy over continuous actions exhibits highly variable lane change times, with lane change manoeuvres that are often too abrupt.

\section{Related Work}\label{sec:related}
\subsection*{Hierarchical and Hybrid Reinforcement Learning}
Various reinforcement learning methods to deal with hierarchical and hybrid action spaces have been proposed, mainly differing in the specific structure of the action space or the considered application. \AuthorCite{Neunert2020} provide a general algorithm for finding optimal hybrid policies with mixed continuous and discrete action spaces. In other works, the focus lies often on hybrid action spaces with a more particular structure, such as parameterized action spaces in PAMDPs~\cite{Masson2016,Fan2019}. This is also the setting considered in this paper for the hybrid options. Additionally, many works on hierarchical reinforcement learning can be interpreted as finding policies over hybrid action spaces, with discrete action selections at the higher levels and continuous action selections at the lower levels (see for example \figref{fig:hierarchical_control}).

An overview of the broader field of hierarchical reinforcement learning is given by \AuthorCite{Pateria2021} and \AuthorCite{Hutsebaut-Buysse2022}. The options (or skills) framework considered in this work is only one specific subfield. The focus in most recent works lies on \emph{learning} a suitable set of options from scratch~\cite{Bacon2017, Zhang2019a}, or learning a policy that can achieve various subgoals~\cite{Kulkarni2016, Levy2019}, or discovering useful skills from large datasets of interaction data~\cite{Pertsch2021, Tanneberg2021}. In contrast, in this work we specifically predefine a small set of options, tailored to the considered autonomous driving tasks, and \emph{apply} them in various hierarchical control setups. From this perspective, our methodology is similar to the works of \AuthorCite{Neumann2009} and \AuthorCite{Dalal2021}, which learn a master policy for selecting the parameters of predefined parameterized skills. Although we apply more recent deep RL techniques and apply intra-option learning, whereas these works and others~\cite{Kulkarni2016, Tessler2017} typically perform Semi-MDP (option-to-option) updates.

\AuthorCite{Barreto2019} and \AuthorCite{Peng2019} introduce algorithms that allow the agent to execute a combination of learned options. Both approaches are more generic than our proposed method for hierarchical control with combined options. More precisely, these methods compose new options acting on the whole (lower level) action space, whereas our approach specifically combines options acting on independent subsets of the action space (dedicated options for longitudinal and lateral control).

\subsection*{Reinforcement Learning for Autonomous Driving}
Our methodology is most comparable with the work of \AuthorCite{Naveed2021} and \AuthorCite{Wang2023}, which apply deep hierarchical RL to learn parameters for dedicated parameterized trajectory planners. A similar strategy is also followed by \AuthorCite{Chen2018}, who first train options for dedicated manoeuvres, and afterwards apply a policy gradient method to learn the master policy over options.
Instead of working with dedicated manoeuvre options, many other works try to learn options with a fixed time horizon from large datasets or from interactions with the environment, which are afterwards used to train a master policy in the Semi-MDP setting~\cite{Zhou2023, Gurses2024, Li2024}.
An important difference with our work is that most of these methods use skills with a fixed time horizon and perform Semi-MDP (option-to-option) updates.
In contrast, our options are predefined and have variable length, while the master policies are trained using intra-option updates. This is crucial to support combined or hybrid options, and allows to seamlessly combine slow lane change manoeuvres with fast acceleration manoeuvres.

\section{Conclusion}
This paper introduced several options tailored to the autonomous driving problem, and showed how they can be applied in various hierarchical control architectures. The resulting framework leads to a very natural representation of the problem, with dedicated options for longitudinal and lateral manoeuvres of varying durations, and separate decision-making modules operating at different timescales and levels in the control hierarchy. In such a setup the learned driving behaviour can also be constrained more easily, as safety and comfort constraints can be individually imposed on the options implementing the manoeuvres. As a result, even though option selection happens at coarser (Semi-MDP) timescales, we can still maintain strong safety guarantees at the finest (MDP) timescale.
Furthermore, in the proposed framework, actions for longitudinal and lateral control can be selected separately, either by options or classical policies, as exemplified by the more flexible policies over combined or hybrid options.
Ultimately, our contributions can lead to more interpretable, reliable, and trustworthy policies for autonomous driving.

Nevertheless, there are still many interesting open directions for future research. For example, proper support for interrupting options could be added. Some smaller experiments with interruption showed frequent abortions of lane change manoeuvres; hence, such an approach should add appropriate deliberation costs on option interruption. Another promising direction for future research would be a more complete implementation of intra-option learning. In our current implementation, the intra-option updates are only applied for active options. Truly updating the value estimates for all options that would take the same action as the active option, could thus further accelerate the learning process. Finally, it would be interesting to investigate parameterizable option policies for generic manoeuvres, such as a lane change manoeuvre with a variable duration parameter, similar to the parameterized trajectories of \AuthorCite{Wang2023}.
Concurrently, additional (more complex) value constraints~\cite{Achiam2017} or dual policy constraints~\cite{DeCooman2024} could be imposed on the driving policies to further improve their performance and reliability.

\subsection*{\small Acknowledgments}
{\small This research was partially supported by Ford, under Ford Alliance Project KUL0076; and TAILOR, a project funded by EU Horizon 2020 research and innovation programme under GA No 952215.
This research received funding from the Flemish Government (AI Research Program). Johan Suykens is also affiliated to Leuven.AI - KU Leuven institute for AI, B-3000, Leuven, Belgium.
The research leading to these results has received funding from the European Research Council under the European Union's Horizon 2020 research and innovation program / ERC Advanced Grant E-DUALITY (787960). This paper reflects only the authors' views and the Union is not liable for any use that may be made of the contained information.
The resources and services used in this work were provided by the VSC (Flemish Supercomputer Center), funded by the Research Foundation - Flanders (FWO) and the Flemish Government.}

\bibliographystyle{abbrvnat}
\bibliography{refs.bib}

\clearpage
\appendices
\setlabelsuffix{app}

\section{Algorithms}\label{app:algorithms}
Below we provide the concrete implementations of the generic \algref{alg:generic}[] for hierarchical reinforcement learning with single options (\algref{alg:opts}), combined options (\algref{alg:combined}) and hybrid options (\algref{alg:hybrid}).

\algblockdefx{With}{EndWith}[1]{\textbf{with}~#1~\textbf{do}}{\textbf{end}}
\algcblock[With]{With}{Else}{EndWith}
\begin{algorithm}
    \caption{Hierarchical RL with Options}\label{alg:opts}
    \algrenewcommand\algorithmicindent{0.9em}
    \begin{algorithmic}
        \State \textbf{Initialize} critic models $\p_{q,i}, \p_{q,i}'$
        \State $\buf \gets \emptyset$
        \For{each episode}
            \For{each environment step $t$}
                \If{$t=0$ or $\s[0] \in \T_{\o[-1]}$} \Comment{Activate new option}
                    \With{probability $\epsilon$} \Comment{$\epsilon$-greedy $\bpol{m}$}
                        \State $\o[0] \sim \uniform{\Oav(\s[0])}$
                    \Else
                        \State $\o[0] = \argmax_{\o \in \Oav(\s[0])} q(\s[0], \o; \p_{q,1})$
                    \EndWith
                \Else
                    \State $\o[0] = \o[-1]$
                \EndIf
                \State $\a[0] = \pi_{\o[0]}(\s)$
                \State $\s[1] \sim \tau(\cdot \given \s[0], \a[0])$
                \State $r_t = r(\s[0], \a[0], \s[1])$
                \State $\buf \gets \buf \union \set{\tuple{\s[0], \o[0], r_t, \s[1]}}$
            \EndFor
            \For{each gradient step}
                \State Sample batch $B$ from $\buf$
                \For{all $\tuple{\s[0], \o[0], r_t, \s[1]} \in B$}
                    \If{$\s[1] \in \T_{\o[0]}$} \Comment{Determine next active option}
                        \State $\o[1] = \argmax_{\o' \in \Oav(\s[1])} q(\s[1], \o'; \p_{q,1})$
                    \Else
                        \State $\o[1] = \o[0]$
                    \EndIf
                    \State $y_t = r_t + \gamma \min_{i \in \set{1, 2}} q(\s[1], \o[1]; \p_{q,i}')$
                \EndFor
                \State $\widehat{L}_q(\p_{q,i}; B) = \frac{1}{\abs{B}}\sum_B \parens[\big]{q(\s[0], \o[0]; \p_{q,i}) - y_t}^2$
                \State $\p_{q,i} \gets \p_{q,i} - \eta_q \nabla_{\p_{q,i}} \widehat{L}_q(\p_{q,i}; B)$
                \State $\p_m' \gets \tau \p_m + (1 - \tau) \p_m'$
            \EndFor
        \EndFor
    \end{algorithmic}
\end{algorithm}

\begin{algorithm}
    \caption{Hierarchical RL with Combined Options}\label{alg:combined}
    \algrenewcommand\algorithmicindent{0.9em}
    \begin{algorithmic}
        \State \textbf{Initialize} critic models $\p_{q,i}, \p_{q,i}'$
        \State $\buf \gets \emptyset$
        \For{each episode}
            \For{each environment step $t$}
                \If{$t=0$ or $\s[0] \in \T_{\o{v}[-1]} \intersec \T_{\o{d}[-1]}$} \Comment{$\epsilon$-greedy $\bpol{m}$}
                    \With{probability $\epsilon$}
                        \State $\tuple{\o{v}[0], \o{d}[0]} \sim \uniform{\Oav{v}(\s[0]) \times \Oav{d}(\s[0])}$
                    \Else
                        \State $\tuple{\o{v}[0], \o{d}[0]} = \argmax\limits_{\o{v} \in \Oav{v}(\s[0]), \o{d} \in \Oav{d}(\s[0])} \mspace{-20mu}q(\s[0], \tuple{\o{v}, \o{d}}; \p_{q,1})$
                    \EndWith
                \ElsIf{$\s[0] \in \T_{\o{v}[-1]}$}
                    \State $\o{d}[0] = \o{d}[-1]$
                    \With{probability $\epsilon$}
                        \State $\o{v}[0] \sim \uniform{\Oav{v}(\s[0])}$
                    \Else
                        \State $\o{v}[0] = \argmax_{\o{v} \in \Oav{v}(\s[0])} q(\s[0], \tuple{\o{v}, \o{d}[0]}; \p_{q,1})$
                    \EndWith
                \ElsIf{$\s[0] \in \T_{\o{d}[-1]}$}
                    \State $\o{v}[0] = \o{v}[-1]$
                    \With{probability $\epsilon$}
                        \State $\o{d}[0] \sim \uniform{\Oav{d}(\s[0])}$
                    \Else
                        \State $\o{d}[0] = \argmax_{\o{d} \in \Oav{d}(\s[0])} q(\s[0], \tuple{\o{v}[0], \o{d}}; \p_{q,1})$
                    \EndWith
                \Else
                    \State $\tuple{\o{v}[0], \o{d}[0]} = \tuple{\o{v}[-1], \o{d}[-1]}$
                \EndIf
                \State $\a[0] = \mat{1 & 0 \\ 0 & 0} \pi_{\o{v}[0]}(\s) + \mat{0 & 0 \\ 0 & 1} \pi_{\o{d}[0]}(\s)$
                \State $\s[1] \sim \tau(\cdot \given \s[0], \a[0])$
                \State $r_t = r(\s[0], \a[0], \s[1])$
                \State $\buf \gets \buf \union \set{\tuple{\s[0], \tuple{\o{v}[0], \o{d}[0]}, r_t, \s[1]}}$
            \EndFor
            \For{each gradient step}
                \State Sample batch $B$ from $\buf$
                \For{all $\tuple{\s[0], \tuple{\o{v}[0], \o{d}[0]}, r_t, \s[1]} \in B$}
                    \If{$\s[1] \in \T_{\o{v}[0]} \intersec \T_{\o{d}[0]}$} \Comment{$\pol{m}$}
                        \State $\begin{gathered}\tuple{\o{v}[1], \o{d}[1]} = \mspace{280mu}\\ \mspace{70mu}\argmax\limits_{\o{v}' \in \Oav{v}(\s[1]), \o{d}' \in \Oav{d}(\s[1])} q(\s[1], \tuple{\o{v}', \o{d}'}; \p_{q,1})\end{gathered}$
                    \ElsIf{$\s[1] \in \T_{\o{v}[0]}$}
                        \State $\o{d}[1] = \o{d}[0]$
                        \State $\o{v}[1] = \argmax_{\o{v}' \in \Oav{v}(\s[1])} q(\s[1], \tuple{\o{v}', \o{d}[1]}; \p_{q,1})$
                    \ElsIf{$\s[1] \in \T_{\o{d}[0]}$}
                        \State $\o{v}[1] = \o{v}[0]$
                        \State $\o{d}[1] = \argmax_{\o{d}' \in \Oav{d}(\s[1])} q(\s[1], \tuple{\o{v}[1], \o{d}'}; \p_{q,1})$
                    \Else
                        \State $\tuple{\o{v}[1], \o{d}[1]} = \tuple{\o{v}[0], \o{d}[0]}$
                    \EndIf
                    \State $y_t = r_t + \gamma \min_{i \in \set{1, 2}} q(\s[1], \tuple{\o{v}[1], \o{d}[1]}; \p_{q,i}')$
                \EndFor
                \State $\widehat{L}_q(\p_{q,i}; B) = \frac{1}{\abs{B}}\sum_B \parens[\big]{q(\s[0], \tuple{\o{v}[0], \o{d}[0]}; \p_{q,i}) - y_t}^2$
                \State $\p_{q,i} \gets \p_{q,i} - \eta_q \nabla_{\p_{q,i}} \widehat{L}_q(\p_{q,i}; B)$
                \State $\p_m' \gets \tau \p_m + (1 - \tau) \p_m'$
            \EndFor
        \EndFor
    \end{algorithmic}
\end{algorithm}

\begin{algorithm}
    \caption{Hierarchical RL with Hybrid Options}\label{alg:hybrid}
    \algrenewcommand\algorithmicindent{0.9em}
    \begin{algorithmic}
        \State \textbf{Initialize} actor and critic models $\p_\mu, \p_\mu', \p_{q,i}, \p_{q,i}'$
        \State $\buf \gets \emptyset$
        \For{each episode}
            \For{each environment step $t$}
                \State $\Delta \tilde{v}_t \sim \gauss_{-1}^{1}(\mu(\s[0]; \p_\mu); \sigma_\epsilon^2)$ \Comment{Truncated Gaussian $\bpol{m,c}$}
                \If{$t=0$ or $\s[0] \in \T_{\o{d}[-1]}$}
                    \With{probability $\epsilon$} \Comment{$\epsilon$-greedy $\bpol{m,d}$}
                        \State $\o{d}[0] \sim \uniform{\Oav{d}(\s[0])}$
                    \Else
                        \State $\o{d}[0] = \argmax_{\o{d} \in \Oav{d}(\s[0])} q(\s[0], \tuple{\Delta \tilde{v}_t, \o{d}}; \p_{q,1})$
                    \EndWith
                \Else
                    \State $\o{d}[0] = \o{d}[-1]$
                \EndIf
                \State $\Delta v_t = \sigma_\mathrm{pwl}(\Delta \tilde{v}_t; \s[0])$ \Comment{State-dependent bounds~\cite{DeCooman2023}}
                \State $\Delta d_t = \mat{0 & 1} \pi_{\o{d}[0]}(\s[0])$
                \State $\a[0] = \mat{\Delta v_t; \Delta d_t}$
                \State $\s[1] \sim \tau(\cdot \given \s[0], \a[0])$
                \State $r_t = r(\s[0], \a[0], \s[1])$
                \State $\buf \gets \buf \union \set{\tuple{\s[0], \tuple{\Delta \tilde{v}_t, \o{d}[0]}, r_t, \s[1]}}$
            \EndFor
            \For{each gradient step}
                \State Sample batch $B$ from $\buf$
                \For{all $\tuple{\s[0], \tuple{\Delta \tilde{v}_t, \o{d}[0]}, r_t, \s[1]} \in B$}
                    \State $\Delta \tilde{v}_{t+1} = \mu(\s[1]; \p_\mu')$ \Comment{$\pol{m,c}'$}
                    \State $\epsilon_\mathrm{c} \sim \gauss(0, \sigma_\mathrm{c}^2)|_{\civ{-c}{c}}$ \Comment{Target policy smoothing~\cite{Fujimoto2018}}
                    \State $\Delta \tilde{v}_{t+1} \gets \Delta \tilde{v}_{t+1} + \epsilon_\mathrm{c}$
                    \If{$\s[1] \in \T_{\o{d}[0]}$} \Comment{$\pol{m,d}$}
                        \State $\o{d}[1] = \argmax\limits_{\o{d}' \in \Oav{d}(\s[1])} q(\s[1], \tuple{\Delta \tilde{v}_{t+1}, \o{d}'}; \p_{q,1})$
                    \Else
                        \State $\o{d}[1] = \o{d}[0]$
                    \EndIf
                    \State $y_t = r_t + \gamma \min_{i \in \set{1, 2}} q(\s[1], \tuple{\Delta \tilde{v}_{t+1}, \o{d}[1]}; \p_{q,i}')$
                \EndFor
                \State $\widehat{L}_q(\p_{q,i}; B) = \frac{1}{\abs{B}}\sum_B \parens[\big]{q(\s[0], \tuple{\Delta \tilde{v}_t, \o{d}[0]}; \p_{q,i}) - y_t}^2$
                \State $\widehat{L}_\mu(\p_\mu; B) = -\frac{1}{\abs{B}}\sum\limits_B \sum\limits_{\o{d} \in \Oav{d}(\s[0])} \mspace{-20mu}q(\s[0], \tuple{\mu(\s[0]; \p_\mu), \o{d}}; \p_{q,1})$
                \State $\widehat{R}_\mathrm{s}(\p_\mu; B) = \frac{1}{\abs{B}}\sum_B \parens[\big]{\mu(\s[1]; \p_\mu) - \Delta \tilde{v}_{t}}^2$
                \State \Comment{Smoothness regularization~\cite{DeCooman2021}}
                \State $\p_{q,i} \gets \p_{q,i} - \eta_q \nabla_{\p_{q,i}} \widehat{L}_q(\p_{q,i}; B)$
                \State $\p_\mu \gets \p_\mu - \eta_\mu \nabla_{\p_\mu} \parens[\big]{\widehat{L}_\mu(\p_\mu; B) + \lambda_\mathrm{s} \widehat{R}_\mathrm{s}(\p_\mu; B)}$
                \State $\p_m' \gets \tau \p_m + (1 - \tau) \p_m'$
            \EndFor
        \EndFor
    \end{algorithmic}
\end{algorithm}

\newpage
\begin{figure}[H]
    \centering
    \includegraphics[width=0.8\linewidth]{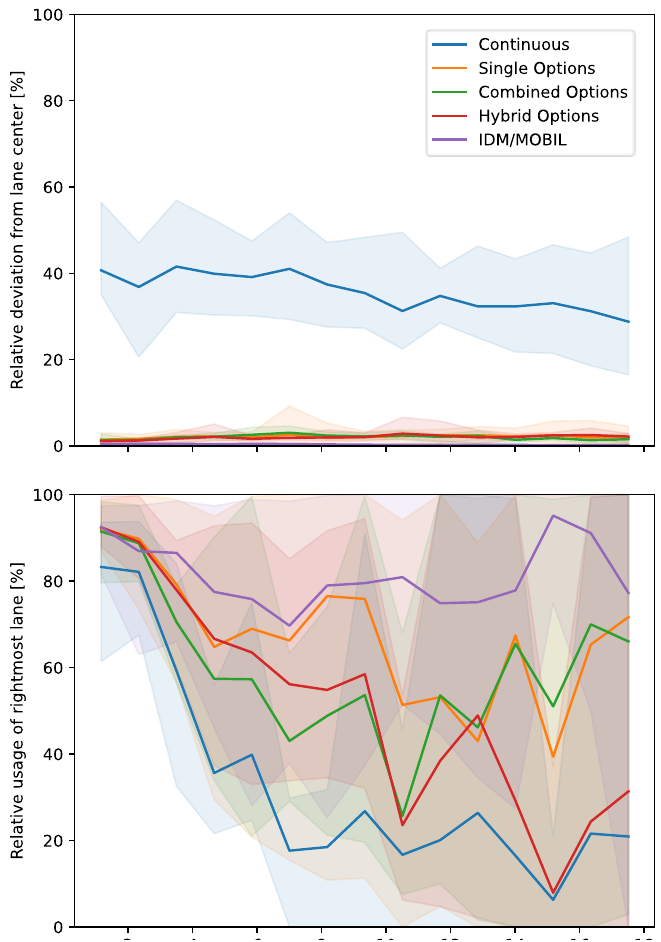} \\[6pt]
    \includegraphics[width=0.8\linewidth]{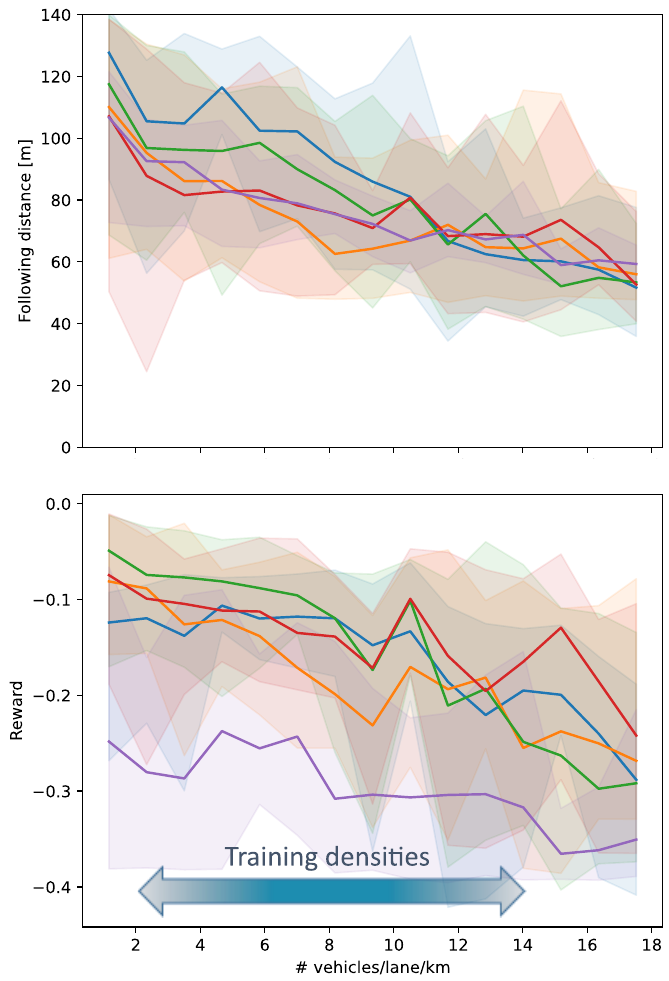}
    \caption{Evaluation of the selected policies in various traffic conditions and comparison with a conservative IDM/MOBIL policy. Solid lines represent the mean value over $10$ episodes, the shaded areas denote the minimum and maximum values.}
    \label{fig:test_extra}
\end{figure}
\begin{figure*}[t]
    \centering
    \includegraphics[width=0.8\linewidth]{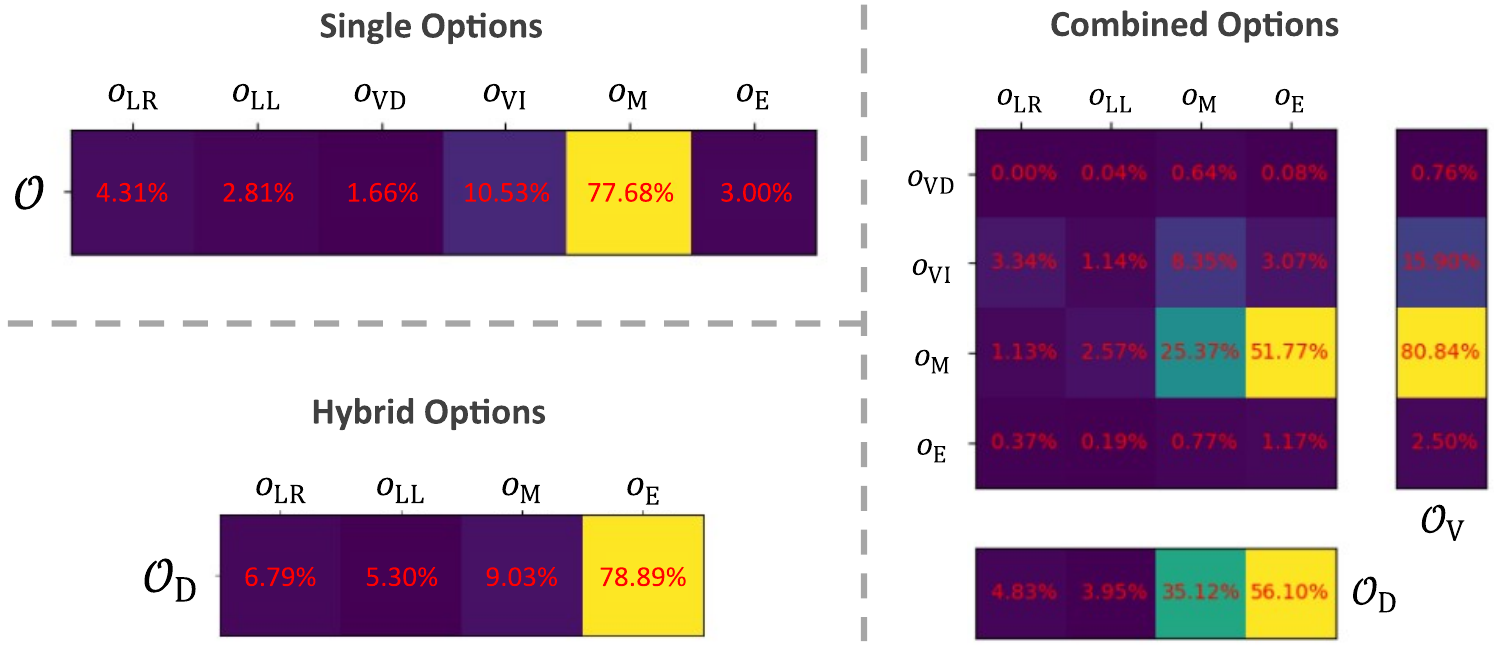}
    \caption{Option activity for the selected policies over options during $E=10$ evaluation episodes. The percentages denote the fraction of time spent with a certain option active.}
    \label{fig:eval_activity}
\end{figure*}
\newpage
\section{Experimental Results}\label{app:results}
Some additional test metrics for the selected policies are shown in \figref{fig:test_extra}, which further supports the observations made in \secref{sec:experiments}[]. In particular, the rewards and following distances of all policies are quite similar and roughly follow the same trends. In more dense traffic conditions, the rightmost (slower) lane is used less frequently by the baseline policy over continuous actions and master policy over hybrid options. This confirms their slight outperformance compared to other approaches in more congested traffic situations. Finally, due to the use of constrained options for lateral control, the hierarchical setups achieve better alignment within the lane with almost no deviation from lane center.

\figref{fig:eval_activity} shows the activity of options during evaluation of the selected (best performing) policies. These plots can be used to analyze the learned driving behaviour in some more detail, and thus increase the interpretability of the learned models as compared to the continuous baseline models.
For example, we can observe that the policy over combined options effectively uses the flexibility of combining longitudinal and lateral manoeuvres, as it actively chooses to alter the vehicle's velocity during lane changes $\sim 50\%$ of the time.
Performing such an analysis or determining which manoeuvres are being performed, is much harder for policies over continuous actions. In fact, only the immediate reference signals selected by a policy over continuous actions are known at any moment in time. In contrast, for the policies over options the exact manoeuvre selected by the master policy is known at every moment in time (and can be determined for any possible traffic situation on the road).

Remark that for lateral control, the maintain and emergency options both provide the same action to maintain the lateral position (unless this would be unsafe). It seems the selected master policies have learned a slight preference for the (lateral) emergency option, which is unintended. This can be easily resolved by adding a small penalty to the reward signal for activating this emergency option. Results for experiments with such a reward modification are however not shown in this work to make a fair comparison between the different approaches easier (using the exact same reward for all experiments).

\section{Hyperparameters}\label{app:hyper}
\tabref{table:hyper} below shows the used hyperparameters for the performed experiments.
\begin{table}[h]
    \centering
	\caption{Overview of the used hyperparameters.}
	\label{table:hyper}
	\begin{tabular}{@{\,} >{\raggedright}m{11em} >{$}l<{$} r @{\;} l r @{\;} l @{\,}}
		\toprule
		\multicolumn{2}{@{\,} l}{\textbf{Hyperparameters}} & \textbf{Values} \\
		\midrule
        Maximum episode length (truncation) & T & \num{5000} \\
        Total training episodes & \, & \num{300} \\
		Warmup timesteps & \, & \num{6400} \\
        Replay buffer size & \abs{\mathcal{B}} & \num{1000000} \\
		Batch size & \abs{B} & \num{64} \\
        Discount factor & \gamma & \num{0.99} \\
        Learning rates (strides) & \eta_q & \num{5e-4} & (1) \\
         & \eta_\mu & \num{5e-4} & (2) \\
		Target networks averaging constant (stride) & \tau & \num{1e-3} & (2) \\
		Network architecture -- & \text{(critic)} & $64 \times 32$ \\
        hidden dimensions & \text{(actor)} & $32 \times 16 \times 8$ \\
		\bottomrule
	\end{tabular}
\end{table}

\end{document}